\documentclass[10pt,twocolumn,letterpaper]{article}

\usepackage{cvpr}              

\usepackage{xcolor}
\usepackage{fontawesome}
\usepackage{cuted}
\usepackage{capt-of}
\usepackage{pifont}
\usepackage{multirow} 
\definecolor{cvprblue}{rgb}{0.21,0.49,0.74}
\usepackage[pagebackref,breaklinks,colorlinks,allcolors=cvprblue]{hyperref}

\def\paperID{11126} 
\def\confName{CVPR}
\def\confYear{2025}

\usepackage{pifont}
\newcommand{\cmark}{\ding{51}}
\newcommand{\xmark}{\ding{55}}

\title{BooW-VTON: \textcolor{red}{Boo}sting In-the-\textcolor{red}{W}ild Virtual Try-On via\\Mask-Free Pseudo Data Training}

\author {
    Xuanpu Zhang\textsuperscript{\rm 1,2}\thanks{Paper work done during an internship at Alibaba.},
    Dan Song\textsuperscript{\rm 1}\thanks{Corresponding author.},
    Pengxin Zhan\textsuperscript{\rm 2},
    Tianyu Chang\textsuperscript{\rm 3},\\ 
    Jianhao Zeng\textsuperscript{\rm 4},
    Qingguo Chen\textsuperscript{\rm 2},
    Weihua Luo\textsuperscript{\rm 2},
    Anan Liu\textsuperscript{\rm 1}{$^{\dagger}$}\\
    \normalsize
$^{1}$\    Tianjin University ~~ $^{2}$\,Alibaba Group ~~ $^{3}$\,University of Science and Technology of China ~~ $^{4}$\,Westlake University\\
}

\begin{document}

\maketitle
\begin{strip}
    \vspace{-2.0cm}
    \centering
    \includegraphics[width=\textwidth]{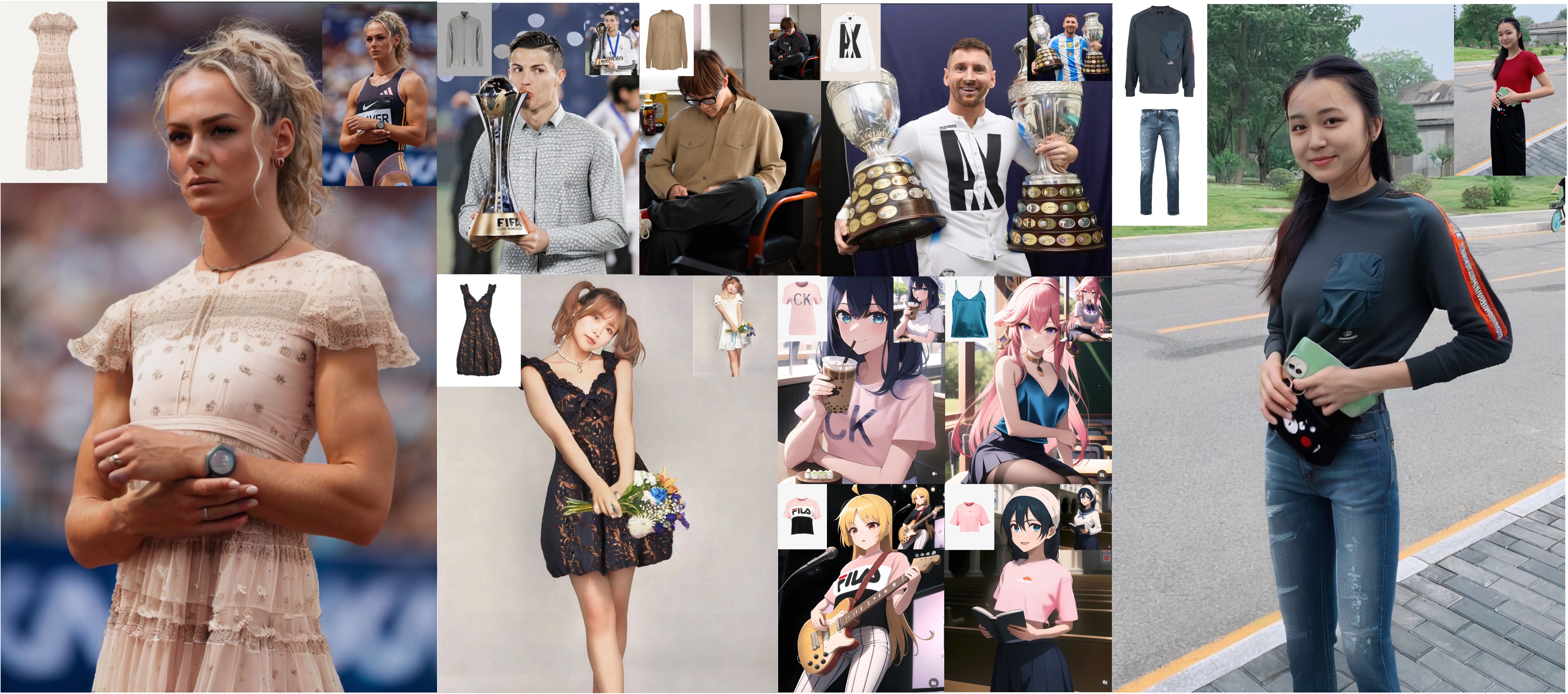}
    \captionof{figure}{Boow-VTON achieves realistic virtual try-on results for in-the-wild scenarios, with complex foreground and background elements. While performing natural try-ons for garments, we faithfully preserve the details of the person (such as accessories, muscles, and skin) and the style of the image (including lighting, shadows, and artistic style).
    \label{fig:main}}
\end{strip}

\begin{abstract}

Image-based virtual try-on is an increasingly popular and important task to generate realistic try-on images of the specific person.
Recent methods model virtual try-on as image mask-inpaint task, which requires masking the person image and results in significant loss of spatial information. Especially, for in-the-wild try-on scenarios with complex poses and occlusions, mask-based methods often introduce noticeable artifacts. Our research found that a mask-free approach can fully leverage spatial and lighting information from the original person image, enabling high-quality virtual try-on. Consequently, we propose a novel training paradigm for a mask-free try-on diffusion model.
We ensure the model's mask-free try-on capability by creating high-quality pseudo-data and further enhance its handling of complex spatial information through effective in-the-wild data augmentation.
Besides, a try-on localization loss is designed to concentrate on try-on area while suppressing garment features in non-try-on areas, ensuring precise rendering of garments and preservation of fore/back-ground.
In the end, we introduce BooW-VTON, the mask-free virtual try-on diffusion model, which delivers SOTA try-on quality without parsing cost.
Extensive qualitative and quantitative experiments have demonstrated superior performance in wild scenarios with such a low-demand input.

\end{abstract}
\vspace{-5mm}
\section{Introduction}
\label{sec:intro}

Image-based virtual try-on (VTON), which aims to generate realistic try-on images of a specific person while preserving the original pose and body feature in source images, 
is an increasingly popular and important task for online shopping.
A key challenge lies in naturally rendering garments onto the correct human body parts while preserving the non-try-on content of the source person image. As shown in Figure \ref{fig:main}, the non-try-on content includes personal characteristics (pose, shape, skin, accessories), scenes (foreground and background), and person image's visual features (light and style).
Based on the powerful diffusion models \cite{DDPM,IDDPM,AutoencoderKL,ControlNet}, current methods \cite{GP-VTON,StableVITON,CAt-DM,OOTD,IDM-VTON,TPD,survey,betterfit} have made substantial progress in simple shop try-on scenarios.
However, complex in-the-wild try-on scenarios reveal defects with these methods.

\begin{figure}[t]
    \centering
    \includegraphics[width=0.48\textwidth]{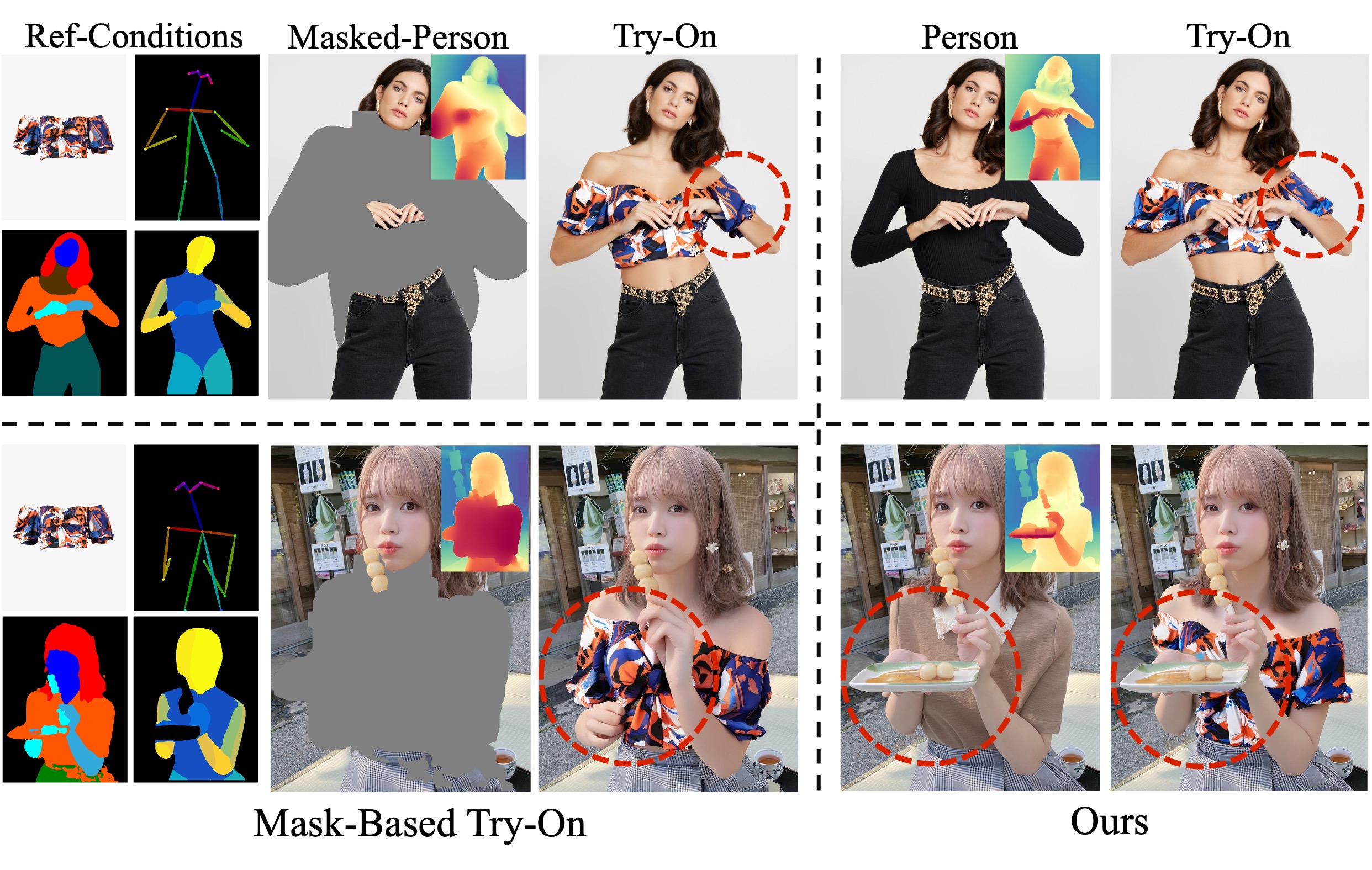}
    \caption{\textbf{Top}: Shop try-on, accurate masks can be obtained through precise human parsing results. The masked person has destroyed spatial information (represented by depth maps). \textbf{Bottom}: In-the-wild try-on, the accuracy of the mask region decreases as the interference of complex scenarios. The masked person suffers from severe loss of spatial information.}
    \label{fig:intro}
\end{figure}

As shown in Figure \ref{fig:intro}, most methods rely on a mask to indicate where to edit and train the model with samples of \{masked person image, target garment image, real person image\}. The mask leads to the loss of person image content and spatial information (represented by the depth map \cite{depth_anything} in Figure \ref{fig:intro}), breaks the coherence of the foreground/background, and requires additional parser to supplement body pose information. These defects are often overlooked in simple shop try-on showcases, where user images typically feature clean portraits. However, in real-world try-on scenarios with complex occlusions and poses, they lead to noticeable artifacts in the generated results.

Recently, some image-based virtual try-on methods have attempted to mitigate the inherent deficiencies of the current try-on pipeline.
For example, TPD \cite{TPD} exploits a more precise mask through a two-stage inference framework that first predicts a more accurate mask and then performs the try-on process with this refined mask. However, this refined operation only considers the shop try-on scenario, it does not resolve the disruption of the person image's information of masked region.
Besides, PFDM \cite{PFDM} introduces a parser-free try-on diffusion model to alleviate the dependence on the mask. Like early GAN-based parser-free methods such as PFAFN \cite{PFAFN}, 
PFDM directly distills a mask-free student model from existing mask-based models.
Unfortunately, although this approach reduces the need for mask in form, its performance is still limited by the constraints of the mask-based model, which cannot generalize to complex in-the-wild try-on cases.

To boost the performance of in-the-wild virtual try-on, we propose a method that fine-tunes a pre-trained latent diffusion model with mask-free pseudo data, enabling it to handle wild scenarios. By leveraging the image understanding capabilities of pre-trained models, this approach adaptively preserves the non-try-on content while rendering the garment.
In a word, we construct high-quality pseudo wild data for strong supervision to teach the model where to edit and concentrate attention on the try-on area to preserve non-try-on content.
Specifically, we first conduct the refined try-on pipeline in simple in-shop scenarios through a powerful mask-based try-on model to generate high-quality pseudo person images.
Then we augment the training pairs \{pseudo person image, real person image\} with diverse background and foreground that are synthesized by a generation model \cite{layerdiffusion}. 
The pseudo wild data is constructed at low cost, providing high-quality guidance to precisely learn clothing-changing area under complex wild interference.
Besides, we design a try-on localization loss to suppress the awareness of garment features in non-try-on areas, ensuring that the garment features are rendered more accurately.
By exploiting such method, we propose BooW-VTON, a powerful mask-free virtual try-on diffusion model. Extensive qualitative and quantitative experiments clearly demonstrate that BooW-VTON significantly outperforms the baseline and other SOTA methods on multiple challenging image-based try-on benchmarks.

Our contributions can be summarized as follows:
\begin{itemize}
    \item We propose BooW-VTON, a mask-free in-the-wild virtual try-on diffusion model that generates realistic try-on results without requiring any additional parser.
    \item We proposed a simple yet effective method to achieve a high-performance try-on model by constructing mask-free pseudo wild data and try-on localization loss.

    \item We verify the superior performance of the proposed method across multiple challenging virtual try-on benchmarks, significantly outperforming the baseline and other state-of-the-art methods.
\end{itemize}

\section{Related Work}
\label{sec:related_work}

\subsection{Mask-Based Virtual Try-On}

Image-based virtual try-on methods can be categorized into two main approaches: mask-based methods and mask-free methods. The primary difference lies in whether a binary mask is required during inference to specify the areas on the person's image where try-on should occur and which areas should remain unchanged.

Mask-based methods \cite{CP-VTON,Viton,VITON-HD,HR-VITON,GP-VTON,ACGPN,CAt-DM,StableVITON,OOTD,IDM-VTON} in virtual try-on treat the task as an image mask-inpainting problem, where the try-on region of a person is inpainted using reference garment. 
Early methods \cite{Viton,CP-VTON,CP-VTON+,ACGPN,VITON-HD,HR-VITON,GP-VTON} utilized dedicated image warping modules to deform garment images to fit the try-on region, guided by human pose parsing \cite{OpenPose,DensePose} features. These modules required accurate human pose features to warp the clothing effectively. With the rapid advancement of diffusion models \cite{IDDPM,DDPM} in image generation, recent methods \cite{OOTD,IDM-VTON,StableVITON,TPD,LaDI-VTON,anyfit} encode garment images and human pose using pre-trained encoders \cite{CLIP,dino}, aligning clothing with human pose in latent space before inpainting within the masked region. However, these approaches suffer from loss of original image content due to masking and heavily rely on the performance of human pose parsers. TPD \cite{TPD} mitigates information loss by reducing the mask area through an additional mask prediction step, but the quality of its predicted masks remains unstable and fails to fully resolve the issues caused by mask.

\subsection{Mask-free Virtual Try-On}

In contrast, mask-free methods \cite{PFAFN,style-flow,WUTON,DCTON} do not require masks as input during inference. They are specifically designed to mitigate the negative effects of parsers. These methods first train a mask-based teacher model, followed by distilling a mask-free student model using results generated from teacher model. 
However, knowledge distillation not only transfers the flaws of the mask-based model to the student network but also limits the student network's generalization ability in wild scenarios. Unlike these methods, we use an optimized inference approach to filter out the flaws of mask-based methods, and then apply data augmentation and loss functions to help the mask-free model acquire a more intrinsic understanding of the try-on concept.

\subsection{In-the-wild Virtual Try-On} 
In-the-wild virtual try-on, designed for consumer scenarios, is gaining increasing attention. However, there are no particularly mature models specifically tailored for wild scenarios. StreetVITON \cite{street_tryon} was an early attempt to address try-on in such scenes but faced challenges with UV mapping \cite{DensePose} for matching body and clothing, resulting in suboptimal try-on effects. Subsequent works like StableVITON \cite{StableVITON} and IDM-VTON \cite{IDM-VTON} leveraged the powerful image restoration capabilities of diffusion models to preserve backgrounds in wild scenarios. However, constrained by mask-based methods, they fail to preserve content within the masked region of the original image, resulting in alterations to user accessories, skin, and other closely adjacent content within the garment region. To enable the model to preserve complex non-try-on content in wild scenes, through pseudo data training, we developed a high-performance mask-free in-the-wild try-on model.

\begin{figure*}[t]
    \centering
    \vspace{-2mm}
    \includegraphics[width=0.96\textwidth]{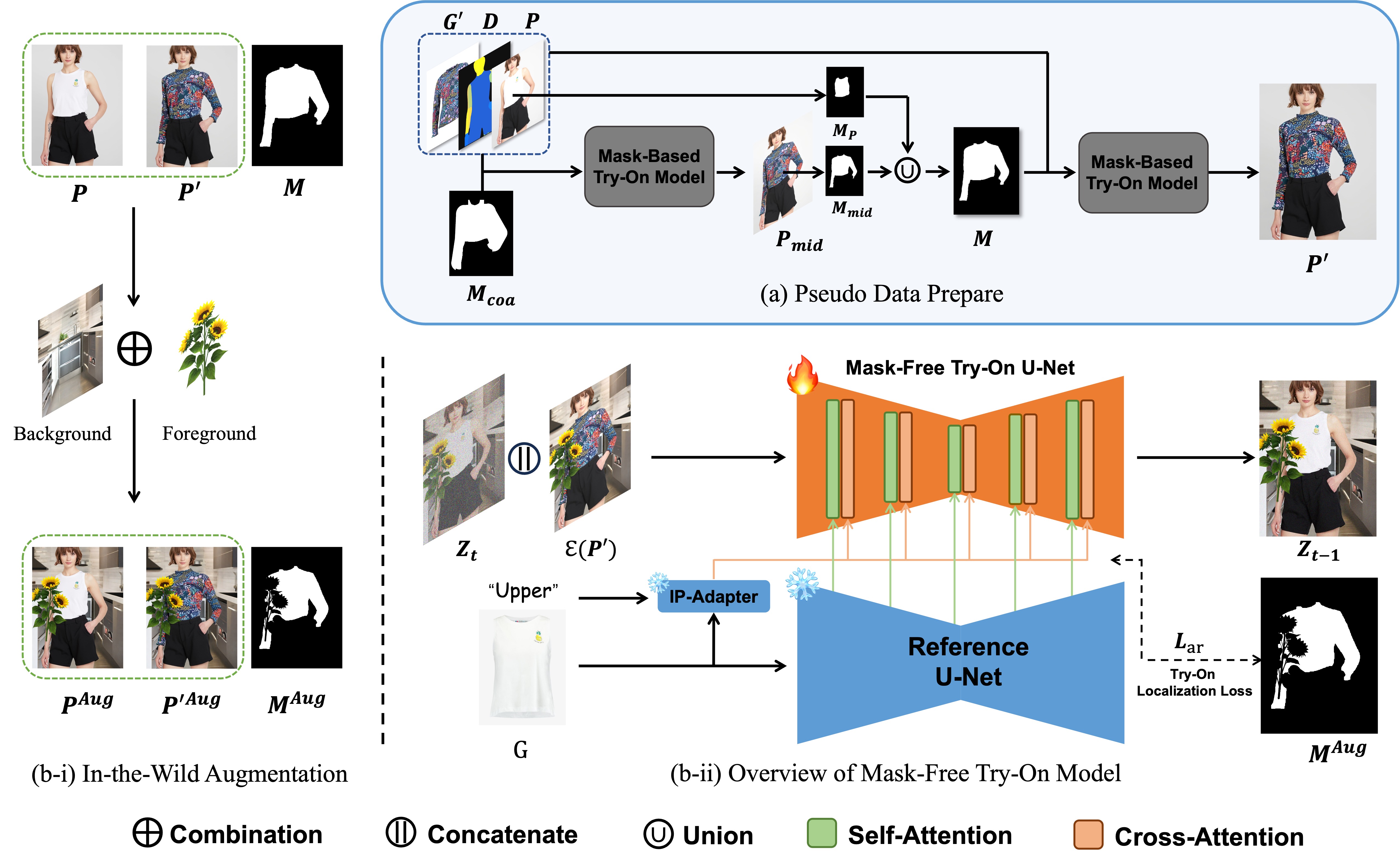}
    \caption{Overview of high-quality in-the-wild try-on model training pipeline. (a) Data prepare for pseudo training pairs. Use the mask-based model with a refined inference pipeline to generate high-quality pseudo-person images. (b-i) Implementation of in-the-wild pseudo pairs. Add background and foreground to the person's image using image stacking. (b-ii) Training process of mask-free try-on diffusion model. Try-on U-Net is trained to determine the try-on regions from the person image $P'$. Use $M^{Aug}$ to constrain the regions for garment alignment and replacement in the attention layer. $M^{Aug}$ is used only during training.}
    \label{fig:method}
    \vspace{-5mm}
\end{figure*}
\section{Preliminary}
\noindent\textbf{Latent Diffusion Model.}
The Latent Diffusion Model \cite{LDM} performs denoising in the latent space of an autoencoder. Image $\mathbf{x}$ is transformed into latent features $\mathbf{z}_0 = \mathcal{E}(\mathbf{x})$ using the autoencoder $\mathcal{E}$.
Given a pre-defined variance schedule $\beta_t$, we can define a forward diffusion process in the latent space according to denoise diffusion probabilistic models~ \cite{DDPM}:
\begin{equation}
    q(\mathbf{z}_t | \mathbf{z}_0) = \mathcal{N}(\mathbf{z}_t ; \sqrt{\bar{\alpha}_t}\mathbf{z}_0, (1-\bar{\alpha}_t)\mathbf{I}),
\end{equation}
where $t \in \{1, ..., T\}$, $T$ represents the number of steps in the forward diffusion process, $\alpha_t:= 1-\beta_t$, and $\bar{\alpha}_t:= \Pi_{s=1}^t \alpha_s$. 
As a training loss, the simplified objective function of latent diffusion model is: 
\begin{equation}
    \mathcal{L}_{LDM} = \mathbb{E}_{\mathcal{E}(\mathbf{x}),y,\epsilon\sim\mathcal{N}(0, 1),t}\left[\lVert\epsilon - \epsilon_{\theta}(\mathbf{z}_t, t, y)\rVert_2^2\right],
    \label{ldm_func}
\end{equation}
where the denoising network $\epsilon_{\theta}(\cdot)$ is implemented with a U-Net architecture~ \cite{UNet} and condition $y$.

\section{Method}
\label{sec:method}

We propose a training pipeline for high-quality in-the-wild mask-free try-on models. First, we use pseudo-triplet data to remove the model's dependency on a mask during training. Then, we enhance the model's try-on capability for complex scenes by leveraging in-the-wild data augmentation. Finally, we introduce a loss function for auxiliary models that effectively learns the try-on regions.

For ease of explanation, we first summarize the variables that may appear in this section:

\begin{itemize}
    \item $P$: a person image;
    \item $P'$: the result of person $P$ wearing garment $G'$;
    \item $P_{mid}$: the middle result of person $P$ wearing garment $G'$;
    \item $D$: densepose map of $P,P',P_{mid}$;
    \item $G$: an in-shop image of the garment worn by the $P$;
    \item $G'$: another garment image different from $G$;
    \item $M_{coa}$: the coarse mask required for trying on $P$.
    \item $M_{P}$: the garment region of $P$.
    \item $M_{mid}$: the garment region of $P_{mid}$.
    \item $M$: the mask required for trying on $P$.
    \item $B$: a clean background image without the subject;
    \item $F$: high-quality foreground transparent image.
\end{itemize}

\subsection{Mask-free Try-On Diffusion Model}
We train the model using pseudo-triplets $\{P', G, P\}$, replacing the original masked-person images with $P'$. This method removes the model’s reliance on masks and reduces mask creation costs. 
Specifically, we use the SDXL \cite{SDXL} as the try-on U-Net, with the pre-trained IP-adapter \cite{ip-adapter} and SDXL-Reference Net \cite{animate_anyone} serving as garment encoders. The garment features encoded by the IP-adapter are injected into the try-on U-Net via cross-attention layers, while the features encoded by the Reference Net are injected through self-attention layers. To maintain the model's extensibility for other tasks, we retained the IP-adapter’s prompt input interface and restricted the prompts to the garment categories \textit{“upper/lower/dresses”}.
We concatenate $Z_t = Noise(\mathcal{E}(P))$ and $\mathcal{E}(P')$ along the channel dimension and input them into the try on U-Net, resulting in the denoised output $Z_{t-1}$. In the attention layer, the model edits the garment content based on the correlation between human body features and garment features.  During training, only the parameters of the try-on U-Net are unfrozen, as shown in Figure \ref{fig:method}-b-ii.

\subsection{High-Quality Pseudo Data Prepare}
To obtain the pseudo-triplets $\{P', G, P\}$ for mask-free model training, we use the mask-based model to generate $P'$ from $P$ and $G'$. In this study, we utilize IDM-VTON \cite{IDM-VTON} as the mask-based model, where the full input is $\{P', D, M, G'\}$, as illustrated in Figure \ref{fig:method}-a. 
The results generated by the mask-based model contain imperfections caused by the mask, which limit the performance of the mask-free model. To address this issue, we adopt a two-stage inference approach in relatively simpler in-shop scenarios to create high-quality pseudo data, as shown in Figure \ref{fig:method}-a. We first use a coarse mask $M_{coa}$ with greater flexibility for redrawing to perform the try-on, obtaining a middle result $P_{mid}$. We extract the garment region $M_{mid}$ from $P_{mid}$, and then perform a union operation with the original person's garment region $M_{P}$ to obtain a more precise mask $M$. With the refine mask $M$ and $\{P', D, G'\}$ we perform a try-on to generate high-quality pseudo-data that retains more non-try-on content.

\subsection{In-the-Wild Data Augmentation}

Since mask-based models are typically trained on shop try-on datasets, the directly obtained pseudo-triplets are usually simple try-on samples. To fully exploit the advantages of the mask-free try-on model and enhance its performance in diverse scenarios, we apply in-the-wild data augmentation to the triplets.

\begin{figure}[t]
    \centering
    \includegraphics[width=0.45\textwidth]{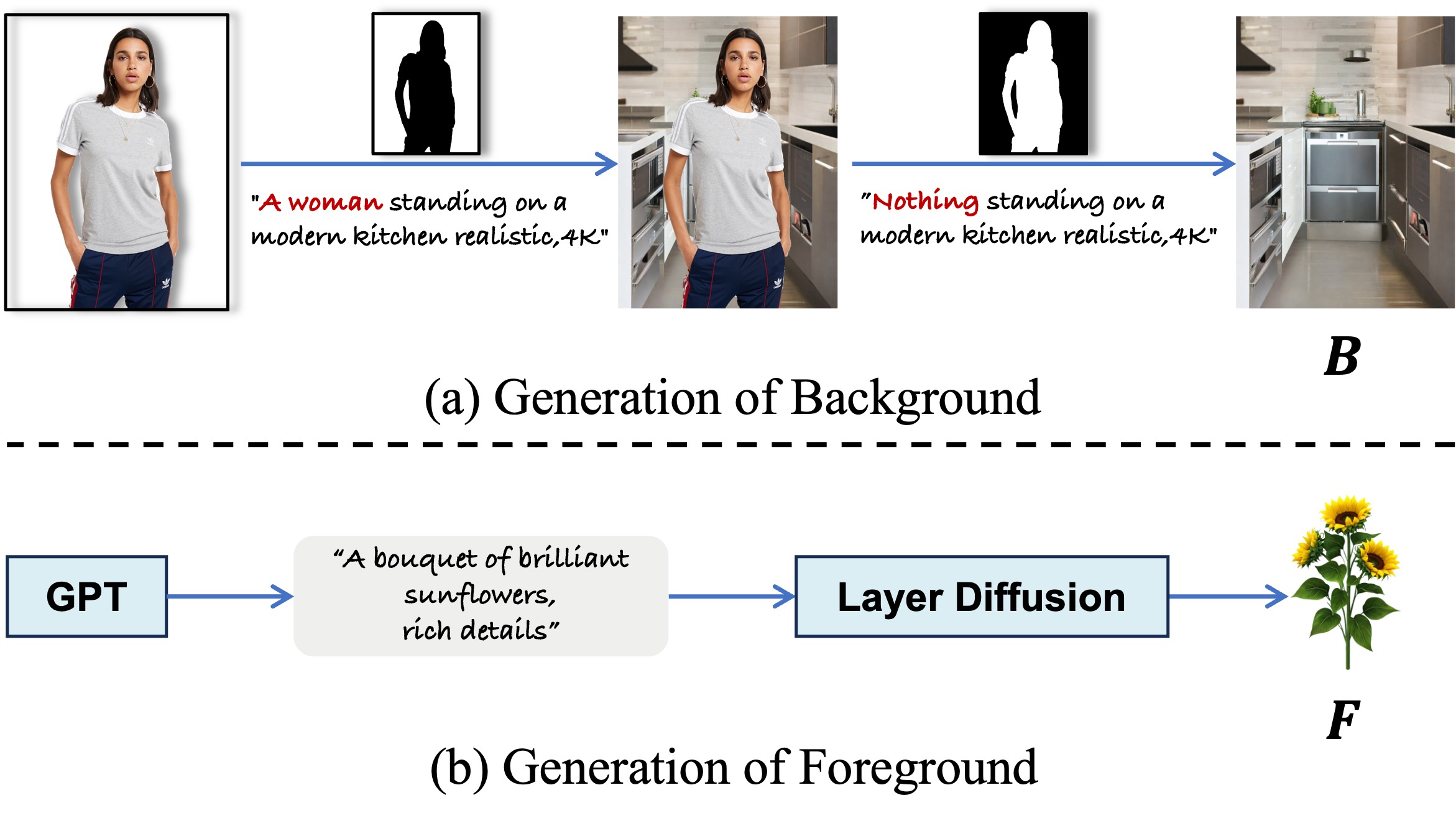}
    \caption{Creating fore/back-ground by synthesizing data.
    \label{fig:fake_data}}
\end{figure}

We use a transparent image of a person and a T2I model \cite{SDXL} to create background $B$, as shown in Figure \ref{fig:fake_data}-a. We first inpaint the blank areas of the person image using the prompt \textit{"A woman \{preposition\} \{scene\}"} to obtain a blended image of the person and background. Then, we modify the prompt template to \textit{"Nothing \{preposition\} \{scene\}"} and inpaint the person's region in the blended image to obtain a clean background image without subject. For the foreground image $F$, as shown in Figure \ref{fig:fake_data}-b, we first obtain a series of prompts for individual objects using GPT-4o. These prompts are then fed into Layerdiffusion \cite{layerdiffusion} to directly generate object images with transparent foregrounds.

During each training iteration, we employ in-the-wild data augmentation on $\{P', P\}$ using the $F$ and $B$. As depicted in Figure \ref{fig:method}-b-i, stacking and combining the images in the order of $B-P/P'-F$ from bottom to top. This placement strategy helps the model differentiate between the foreground and the try-on regions while avoiding excessive occlusion that could completely obscure the garment pixels. The occluded regions created by the foreground should be non-try-on areas. Therefore, we modify the try-on mask $M$ to be used for regularization in these non-try-on areas. In the implementation process, we randomly select foreground and background images and apply translation, compositional, and scaling transformations to achieve better data augmentation results.

\begin{figure}[t]
    \centering
    \includegraphics[width=0.45\textwidth]{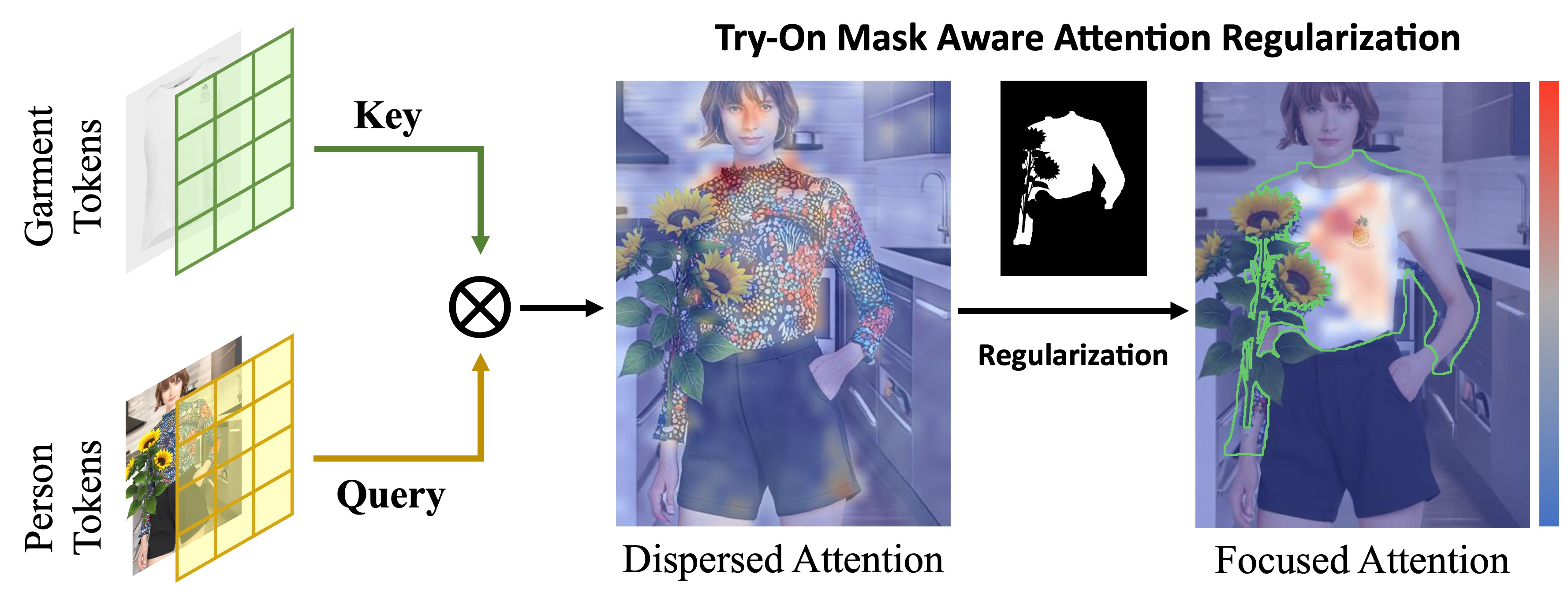}
    \caption{Using try-on mask $M^{Aug}$ to guide the model’s attention to the correct try-on areas.
    \label{fig:attn_reg}}
\end{figure}
\vspace{-1mm}
\subsection{Try-On Localization Loss}
To help the model correctly identify try-on areas and edit content within these areas while preserving the content in non-try-on regions, We apply attention regularization in attention layers as the try-on localization loss.

The attention layer accepts the person latent code $p \in \mathbb{R}^{(h \times w) \times f}$ and the garment tokens $g \in \mathbb{R}^{l \times f}$ as inputs. It then projects the latent code and garment tokens into Query, Key, and Value matrices: $Q_p = W^q p$, $K_g = W^k g$, and $V_g = W^v g$. Here, $W^q, W^k, W^v \in \mathbb{R}^{f \times d}$, represent the weight matrices of the three linear layers, and $d$ is the dimension of Query, Key, and Value embeddings.
The attention layer then computes the attention scores $A = \text{Softmax}(\frac{Q_{p}K_{g}^T}{\sqrt{d}}) \in [0,1]^{(h \times w) \times n}$, and takes a weighted sum over the Value matrix to obtain the attention layer output $p_\text{attn} = AV_g \in \mathbb{R}^{(h \times w) \times d}$. Intuitively, the attention layer scatters garment content to the 2D person latent code space, and $\mathcal{A} = \{A_1, A_2, \dots A_n\}$ represents the amount of information flow from the $k$-th garment token to the $(i, j)$ latent pixel. As shown in Figure \ref{fig:attn_reg}, the attention map should focus solely on the try-on area rather than spreading throughout the entire image, preventing changes in the non-try-on area. To accomplish this, we constrain the attention scores in non-try-on areas using try-on mask $M^{Aug}$. Let $A_{k} = A[:,:,k] \in [0,1]^{(h \times w)}$ be the attention map of the $k$-th garment token. We use the complement of the try-on mask to represent the non-try-on area, minimizing attention scores within this region:
\begin{equation}
    \mathcal{L}_{\text{ar}} = \frac{1}{n}\sum_{k=1}^n \text{mean}(A_k(1-M^{Aug})).
\end{equation}

Employing LDM loss across all regions to ensure accurate garment deformation and smooth consistency of image content:

\begin{equation}\label{eq:ldm_loss}
    \mathcal{L}_{LDM} = \mathbb{E}_{\mathbf{z}_t,\zeta,g,\epsilon,t}\left[\lVert\epsilon - \epsilon_{\theta}(\mathbf{z_t}, \zeta,t, g)\rVert_2^2\right],
\end{equation}
where $\zeta = \mathcal{E}(P')$. The final loss function is:
\begin{equation}
    \mathcal{L} = \mathcal{L}_{\text{LDM}} + \lambda_{\text{ar}} \mathcal{L}_{\text{ar}},
\end{equation}
where $\lambda_{\text{ar}}$ is weight hyper-parameter.

\section{Experiments}

\noindent \textbf{Datasets.} 
We trained our models separately on the VITON-HD \cite{VITON-HD} and DressCode \cite{DressCode} datasets. We perform cross-dataset validation of the model's performance on the in-the-wild datasets StreetVTON \cite{street_tryon} and WildVTON. For more details about the dataset, please refer to Table \ref{tab:comparsion_dataset}.

\begin{table}[h]
\centering
\footnotesize
\label{tab:comparison_data_set}
\setlength{\tabcolsep}{.35em}
\resizebox{\linewidth}{!}{
\begin{tabular}{l c ccccc}
\toprule
\textbf{Dataset} & & \textbf{Images}   & \textbf{Resolution}& \textbf{Pairs} & \textbf{Wild Back} & \textbf{Wild Fore} \\
\midrule
VITON-HD         & & $2032$            & $1024\times768$    & \cmark         & \xmark                   & \xmark                   \\
DressCode        & & $1800 \times 3$   & $1024\times768$    & \cmark         & \xmark                   & \xmark                   \\
StreetVTON       & & $2089$            & Various            & \xmark         & \cmark                   & \xmark                   \\
WildVTON         & & $1224$             & Various            & \xmark         & \cmark                   & \cmark                   \\
\bottomrule
\end{tabular}
}
\caption{Comparison between different test dataset.}
\label{tab:comparsion_dataset}
\end{table}

\begin{figure*}[t]
    \centering
    \includegraphics[width=\textwidth]{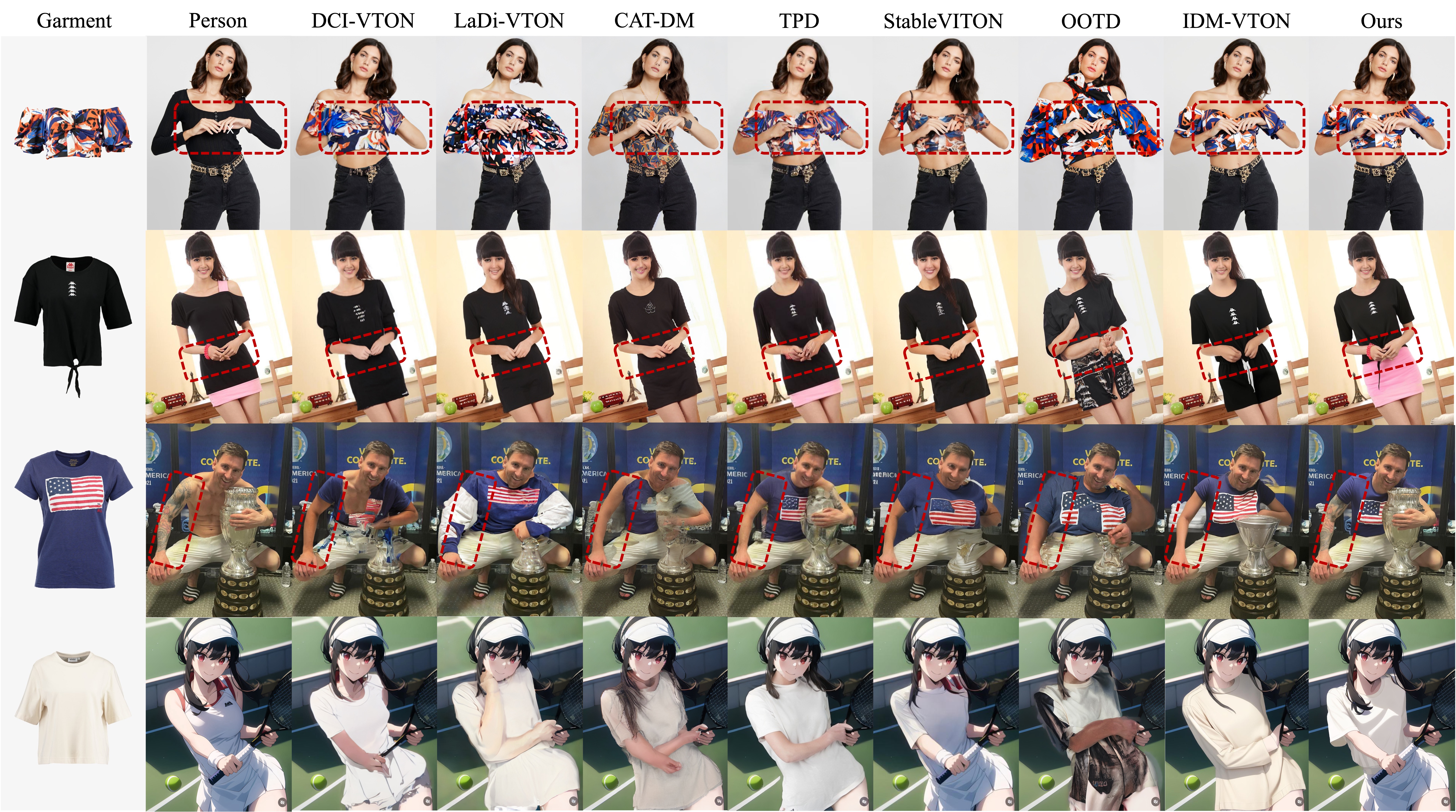}
    \caption{Qualitative comparison of VITON-HD/StreetVITON/WildVTON/Anime person try-on using VITON-HD garment.
    \label{fig:comparsion_hd}}
\end{figure*}

\noindent \textbf{Baselines.} 
We compare our method with previous virtual try-on methods, including DCI-VTON \cite{DCI-VTON}, LaDi-VTON \cite{LaDI-VTON}, CAT-DM \cite{CAt-DM}, StableVITON \cite{StableVITON}, TPD \cite{TPD}, OOTD \cite{OOTD}, and IDM-VTON \cite{IDM-VTON}. We employ pre-trained checkpoints provided in official repositories for testing. 

\noindent\textbf{Evaluation Detail.} 
We perform paired and unpaired testing on the VITON-HD and DressCode test sets, and unpaired testing on the StreetVTON and WildVTON test sets. For paired testing, we generate pseudo pairs for the test set using the same strategy as during training. LPIPS \cite{LPIPS}, SSIM \cite{SSIM} and PSNR are used as paired evaluation metrics. FID \cite{FID} and KID \cite{KID} are used as unpaired metrics. To eliminate the impact of resolution on the metric values, we follow IDM-VTON's approach by standardizing all results to a resolution of $1024 \times 768$ for testing.

\noindent \textbf{Implementation Details.} 
We use Diffusers for implementing our training code and initialize the model using official weights from IDM-VTON \cite{IDM-VTON}. We train our models for about 12 hours using 16 NVIDIA H100 GPUs. We set the learning rate to 5e-6, utilized the Adam optimizer, and trained with a batch size of 32 for 12,000 steps. The weight hyper-parameters were set to $\lambda_{\text{ar}} = 1$. For memory efficiency, we apply the try-on localization loss to attention blocks 5 through 64 of the 70 attention blocks in SDXL, where the image tokens length within these blocks is $32\times24$. And the try-on localization loss is applied separately to the cross-attention and self-attention layers within each of these attention blocks. We use an RTX 4090 GPU to infer our models with 30 DDIM \cite{DDIM} sampling steps.

\begin{figure}[t]
    \centering
    \includegraphics[width=0.48\textwidth]{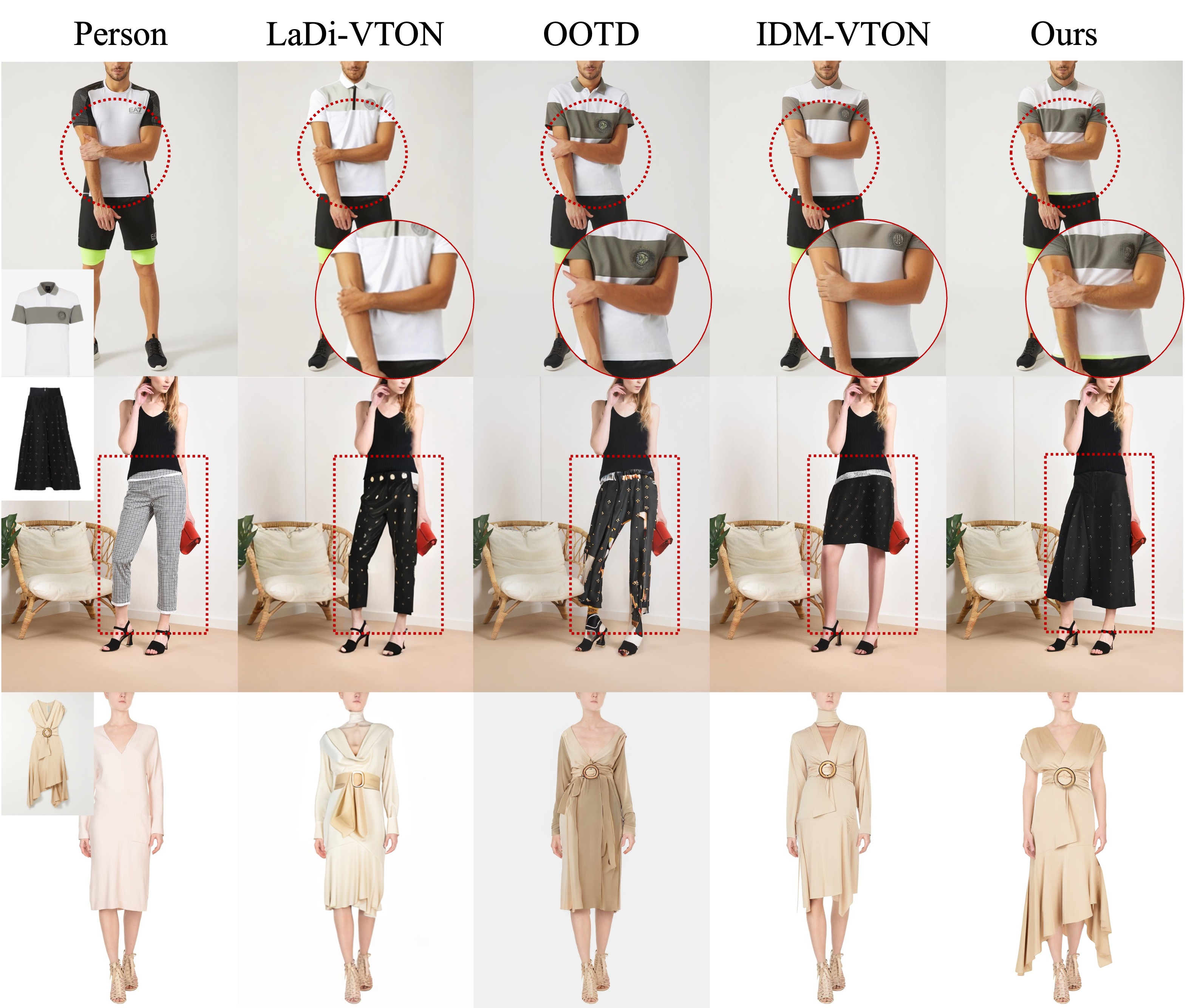}
    \caption{Qualitative comparison of DressCode.
    \label{fig:comparsion_dc}}
\end{figure}

\subsection{Qualitative Results}

\begin{table*}[t]
\centering
\footnotesize
\setlength{\tabcolsep}{.3em}
\resizebox{0.9\linewidth}{!}{
\begin{tabular}{lcc ccccccc c cc c cc}
\toprule
\textbf{Testsets} &&& \multicolumn{7}{c}{\textbf{VITON-HD}} && \multicolumn{2}{c}{\textbf{StreetVTON}}&&\multicolumn{2}{c}{\textbf{WildVTON}}\\
\cmidrule{4-10} \cmidrule{12-13} \cmidrule{15-16}
\textbf{Model} 
&&& \textbf{LPIPS} $\downarrow$ & \textbf{SSIM} $\uparrow$ & \textbf{PSNR} $\uparrow$ & $\textbf{FID}_\text{p} \downarrow$ & $\textbf{KID}_\text{p} \downarrow$ & $\textbf{FID}_\text{u} \downarrow$ & $\textbf{KID}_\text{u} \downarrow$ && $\textbf{FID}_\text{u} \downarrow$ & $\textbf{KID}_\text{u} \downarrow$  && $\textbf{FID}_\text{u} \downarrow$ & $\textbf{KID}_\text{u} \downarrow$ \\
\midrule
DCI-VTON          &MM2023             & & 0.1800 & 0.8545 & 19.27 & 24.05 & 16.44 & \underline{
8.998} & 1.187 && 20.95 & \textbf{3.470}  && 35.66 & \underline{4.649}  \\ 
LaDI-VTON         &MM2023             & & 0.2014 & 0.8395 & 18.69 & 9.746 & 2.599 & 11.08 & 2.634 && 24.12 & 5.638  && 44.54 & 9.203  \\ 
CAT-DM            &CVPR2024             & & 0.1621 & 0.8391 & 20.45 & 9.336 & 2.294 & 10.28 & 1.980 && 25.84 & 6.879  && 42.16 & 8.374  \\ 
TPD               &CVPR2024             & & 0.1822 & 0.8516 & 20.75 & 13.07 & 7.880 & 13.82 & 6.641 && 23.02 & 4.671  && 45.37 & 14.67  \\ 
StableVITON       &CVPR2024             & & 0.1479 & 0.8519 & \underline{21.72} & 8.926 & 2.538 & 9.851 & 1.727 && 23.15 & 4.628  && 42.32 & 8.194  \\ 
OOTD              &\faGithub5.9k stars             & & 0.1420 & 0.8301 & 19.20 & 8.136 & 1.469 & 12.19 & 2.682 && 27.00 & 7.473  && 40.68 & 5.606  \\ 
IDM-VTON          &ECCV2024             & & 0.1223 & 0.8547 & 21.06 & 8.594 & 2.529 & 9.265 & 1.272 && 23.62 & 6.181  && 38.77 & 7.686  \\  \hline
Base mask-free         &            & & 0.1206 & 0.8529 & 20.60 & 8.766 & 2.611 & 9.467 & 1.493 && 28.81 & 8.026  && 57.52 & 18.59  \\ 
+H.Q. pseudo data     &         & & 0.1101 & \underline{0.8597} & 21.37 & \underline{6.896} & 1.580 & 9.191 & 1.120 && 27.26 & 7.616  && 56.14 & 18.31  \\ 
+Wild augmentation             &         & & \underline{0.1173} & 0.8589 & 21.23 & 7.405 & \underline{1.405}  & 9.204 & \underline{1.089}  && \underline{21.70} & 5.170 && \underline{35.62} & 6.180   \\ 
+$L_\text{ar}$ (Full Model)    &  & & \textbf{0.1080} & \textbf{0.8618} & \textbf{21.80} & \textbf{6.885} & \textbf{1.366} & \textbf{8.809} & \textbf{0.8176} && \textbf{20.50} & \underline{4.494} && \textbf{32.53}  & \textbf{4.509}     \\ 
\bottomrule
\end{tabular}
}
\caption{Quantitative results on the VITON-HD/StreetVTON/WildVTON test set with training on the VITON-HD dataset.}
\label{tab:comparison_wild}
\end{table*}

Compared to existing methods, BooW-VTON excels in various try-on scenarios. In the first row of Figure \ref{fig:comparsion_hd}, we show try-on results for complex poses from VITON-HD, where arm occlusions create challenges for existing methods. Our approach successfully renders the clothing while preserving finger and arm details, significantly outperforming existing models. The second row demonstrates our method's ability to maintain garment fit and accessories using the StreetVTON dataset. Unlike existing methods, which are affected by the original garment, our method accurately adjusts for the differences between the original and target garments, filling in missing areas and preserving finger and accessory details for a more realistic result. In the third row, we highlight BooW-VTON's ability to preserve tattoos and foreground details, achieving more accurate clothing rendering while retaining original image content. Finally, we show that even without training on anime-style samples, our method effectively preserves garment features and aligns with the original style.

In Figure \ref{fig:comparsion_dc}, we show BooW-VTON's performance on the DressCode dataset. In the first row, our method preserves muscle details and enhances realism by rendering clothing wrinkles due to compression. In the second row, BooW-VTON accurately reproduces the skirt's features, unaffected by the original pants' fit. The third row demonstrates our method's ability to preserve detailed fabric textures in the dress.

\subsection{Quantitative Results}

As shown in Tables \ref{tab:comparison_wild} and \ref{tab:comparison_dresscode}, we compare the quantitative metrics of BooW-VTON with existing methods across four datasets: VITON-HD, DressCode, StreetVTON, and WildVTON. Thanks to its excellent skin preservation and clothing rendering capabilities, BooW-VTON outperforms existing methods across all metrics on the VITON-HD and DressCode datasets. For in-the-wild try-on scenarios from StreetVTON and WildVTON, our method shows significant advantages over most existing approaches. DCI-VTON performs better than BooW-VTON on the KID metric in StreetVTON, as it leverages a mask to enhance the preservation of non-try-on region pixels. However, this approach overlooks the consistency of the overall image, leading to inferior visual try-on results compared to our method.

\begin{table}[t]
\centering
\footnotesize
\setlength{\tabcolsep}{.3em}
\vspace{2mm}
\resizebox{\linewidth}{!}{
\begin{tabular}{lc ccccc c cc}
\toprule

\textbf{Model} 
& &  \textbf{LPIPS} $\downarrow$ & \textbf{SSIM} $\uparrow$ & \textbf{PSNR} $\uparrow$ & $\textbf{FID}_\text{p} \downarrow$ & $\textbf{KID}_\text{p} \downarrow$ & & $\textbf{FID}_\text{u} \downarrow$ & $\textbf{KID}_\text{u} \downarrow$ \\
\midrule
\textbf{Upper}     & &         &        &       &       &       & &        &       \\
LaDI-VITON         & &  0.1091 & 0.9044 & 21.17 & 18.14 & 3.703 & & 16.43  & 4.829 \\

OOTD               & &  0.0855 & 0.8997 & 20.74 & 16.20 & 5.862 & & 13.20 & 1.860 \\
IDM-VTON           & &  0.0761 & 0.9125 & 22.98 & 16.25 & 7.352 & & 13.60 & 2.952 \\ \hline
Ours               & &  \textbf{0.0615} & \textbf{0.9187} & \textbf{23.81} & \textbf{8.941} & \textbf{1.518} & & \textbf{11.03} & \textbf{0.8618} \\
\midrule
\textbf{Lower} && & & & & && & \\
LaDI-VITON         & &  0.1314 & 0.8855 & 20.50 & 14.98  & 3.920 & & 13.95  & 2.564 \\

OOTD               & &  0.1168 & 0.8706 & 19.09 & 15.56  & 3.797 & & 21.50  & 8.248 \\
IDM-VTON           & &  0.1103 & 0.8869 & 20.45 & 18.20 & 8.123 & & 15.97 & 5.386 \\ \hline
Ours               & &  \textbf{0.0831} & \textbf{0.8970} & \textbf{22.14} & \textbf{10.67} & \textbf{2.344} & & \textbf{13.74} & \textbf{2.455} \\
\midrule
\textbf{Dresses} && & & & & && & \\
LaDI-VITON         & &  0.1753 & 0.8424 & 18.30 & 24.00  & 13.70  & & 16.86  & 5.005 \\

OOTD               & &  0.1490 & 0.8440 & 17.24 & 25.75  & 15.29 & & 20.95  & 8.149 \\
IDM-VTON           & &  0.1381 & 0.8627 & 18.85 & 25.96  & 17.67 & & 12.82 & 9.739 \\ \hline
Ours               & &  \textbf{0.1044} & \textbf{0.8741} & \textbf{20.95} & \textbf{8.434}  & \textbf{1.012}  & & \textbf{10.23} & \textbf{0.5599} \\

\bottomrule
\end{tabular}
}
\caption{Quantitative comparison results on DressCode.}
\label{tab:comparison_dresscode}
\end{table}

\subsection{Ablation Study} \label{sec:ablation}

\begin{figure*}[t]
    \centering
    \includegraphics[width=\textwidth]{images/ablation_study.jpg}
    \caption{\textbf{Left:} Visualization comparison of ablation study. \textbf{Right:} Visualization of attention map according to try-on localization loss.
    \label{fig:ablation_attn_vis}}
\end{figure*}

\noindent\textbf{Ablation Study of Mask-Free Pseudo Data Training.} We conduct ablation experiments on the VITON-HD, StreetVTON, and WildVTON datasets. Starting with a traditional mask-free method, we distill knowledge from IDM-VTON to establish a baseline, then progressively apply our improvements. All experiments use the same training strategy. 

As shown in Table \ref{tab:comparison_wild}, converting a mask-based model to a mask-free model does not improve performance in in-shop try-on scenarios and worsens performance in wild scenarios. Figure \ref{fig:ablation_attn_vis}-Left shows that simple distillation leads to overfitting in in-shop try-on tasks, significantly reducing model generalization. High-quality pseudo data (H.Q.) improves in-shop performance, but the model still struggles in wild scenarios. After applying in-the-wild data augmentation, the model's try-on ability in wild scenarios improves significantly. As shown in Table \ref{tab:comparison_wild}, both FID and KID scores decrease on the StreetVTON and WildVTON test sets. However, there is a slight drop in in-shop performance, likely due to slower convergence from strong data augmentation. Figure \ref{fig:ablation_attn_vis}-Left shows that non-try-on content is well-preserved. Adding try-on localization loss to regularize the attention layers significantly improves performance, reducing visual artifacts and better preserving foreground details, as shown in the red box of Figure \ref{fig:ablation_attn_vis}-Left.

\noindent\textbf{Effectiveness of Try-on Localization Loss.} We analyze the effectiveness of the try-on localization loss by visualizing the self-attention layer of the 35th attention block. Following the diffusion model's efficiency analysis \cite{ediff,dift}, we focus on self-attention maps at steps 1, 15, and 30 during DDIM denoising with 30 steps. As shown in Figure \ref{fig:ablation_attn_vis}-Right, after applying try-on localization loss, the attention becomes more focused on the try-on regions, improving the preservation of subtle foreground details.

\subsection{Application} \label{sec:application}

\begin{figure}[b]
    \centering
    \vspace{-4mm}
    \includegraphics[width=0.48\textwidth]{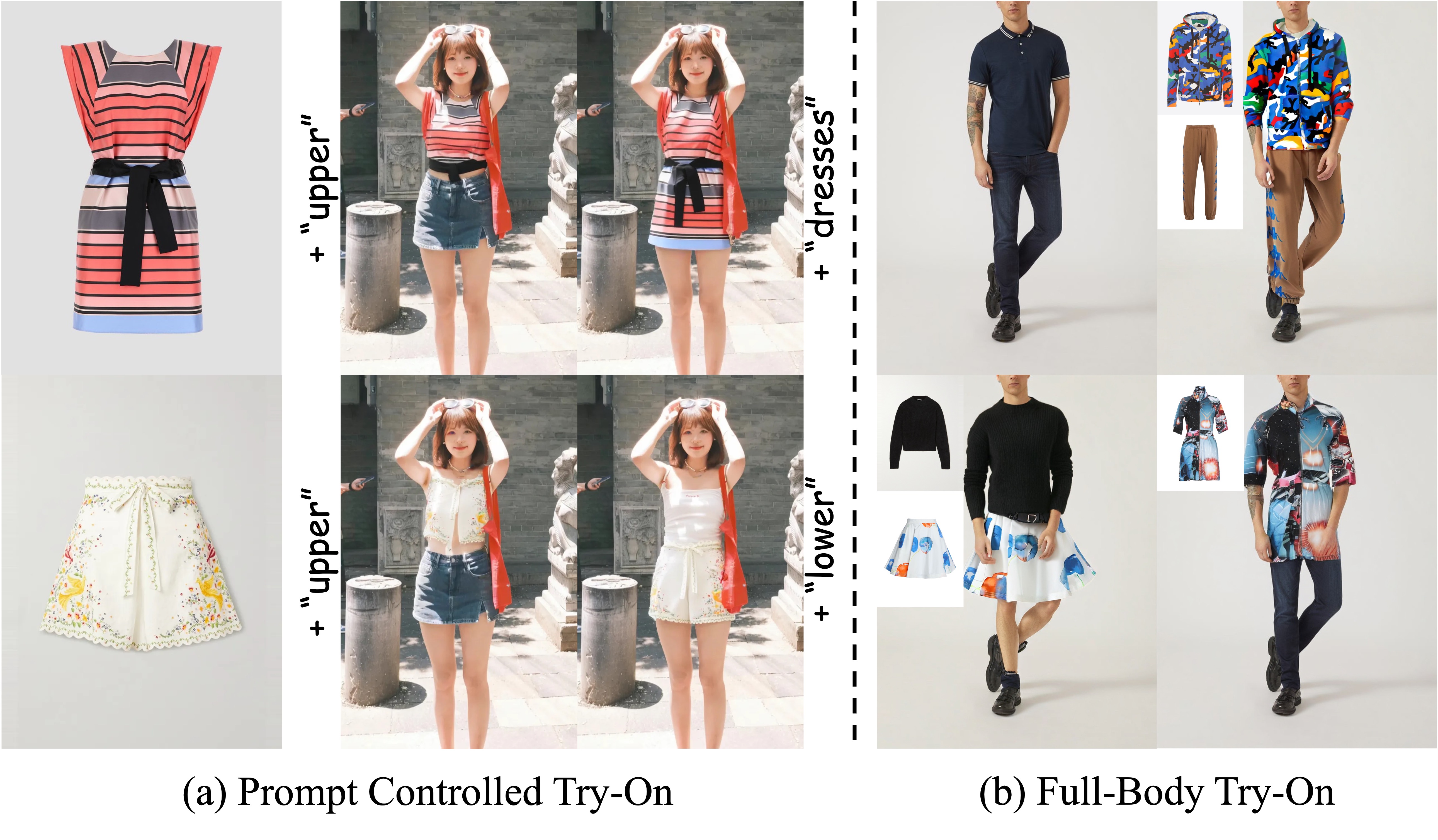}
    \vspace{-4mm}
    \caption{\textbf{Left:} Visual results of different prompts and garment combinations. \textbf{Right:} Visual results of full-body try-on.
    \vspace{-4mm}
    \label{fig:application}}
\end{figure}

Our method enables diverse, high-quality virtual try-on applications. As shown in Figure \ref{fig:application}-Left, the try-on style can be controlled for specified garment items by adjusting the prompt, allowing for the generation of varied garment renderings useful in fashion design. As illustrated in  Figure \ref{fig:application}-Right, even without any full-body try-on training, our method can still produce naturally, realistic full-body try-on results.

\begin{figure}[t]
    \centering
    \includegraphics[width=0.47\textwidth]{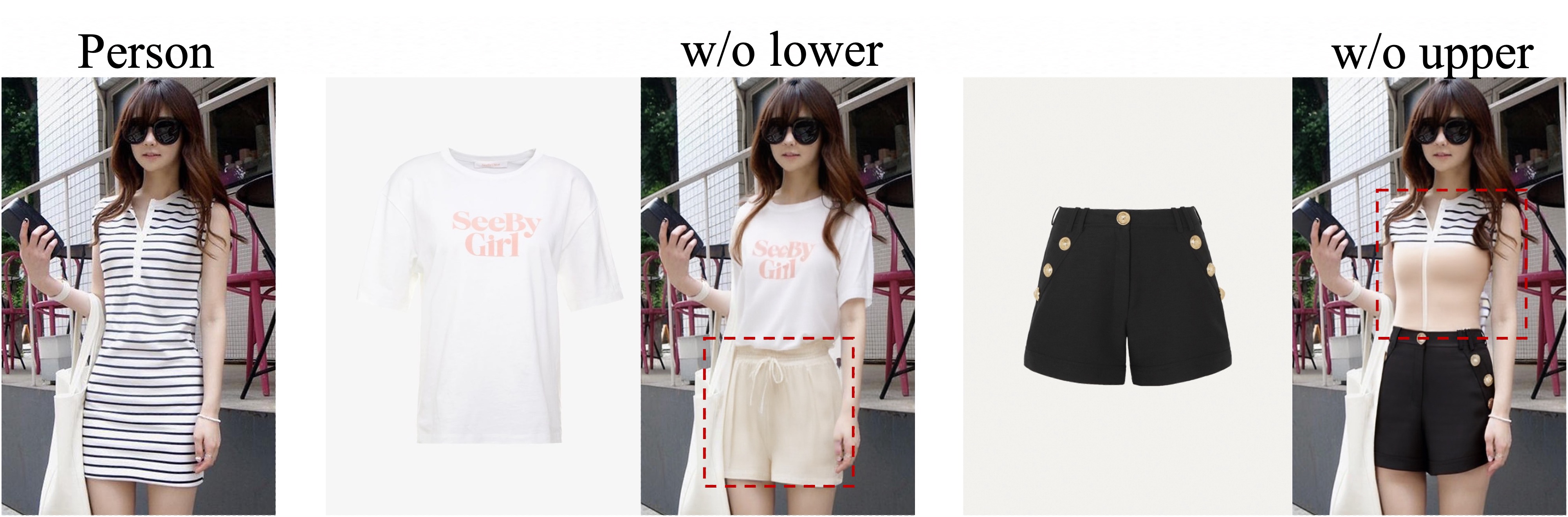}
    \caption{Absence of reference garment brings random content.
    \vspace{-1mm}
    \label{fig:limitation}}
    \vspace{-4mm}
\end{figure}

\section{Limitations}

While our method extends virtual try-on to more general wild scenarios, it still has limitations in user-controllability, which limits its consumer application. As shown in Figure \ref{fig:limitation}, when a lady wearing a dress tries on a T-shirt, the lower of person is randomly generated due to the lack of reference for the lower garment, which may not align with the desired outfit style. A similar issue also occurs in the try-on of lower garments. Therefore, enabling users to achieve their desired outfits, including matching upper and lower garments or accessories, is a key challenge for future research.

\vspace{-2mm}
\section{Conclusion}
In this paper, we propose a novel image-based try-on training method to eliminate the damage of try-on mask in the current learning framework, thereby achieving more superior and realistic in-the-wild try-on performance compared to the existing state-of-the-art methods.  
Specifically, the person image pairs with only different garment are constructed as our training data of our mask-free try-on model.
Besides, due to the difficulty for capturing accurate try-on area in more complex wild scenarios,
we introduce the in-the-wild data augmentation and try-on localization loss to further boost performance in real-world try-on scenarios. 
Extensive qualitative and quantitative experiments demonstrate that our model surpasses existing methods across diverse try-on situations.

{
    \small
    \bibliographystyle{ieeenat_fullname}
    \bibliography{main}
}


\clearpage
\setcounter{section}{0}
\maketitlesupplementary

\section{Method Details}
\label{wild_aug}

\subsection{High Quality Pseudo Data Details}

\begin{figure*}[h]
    \centering
    \includegraphics[width=1\textwidth]{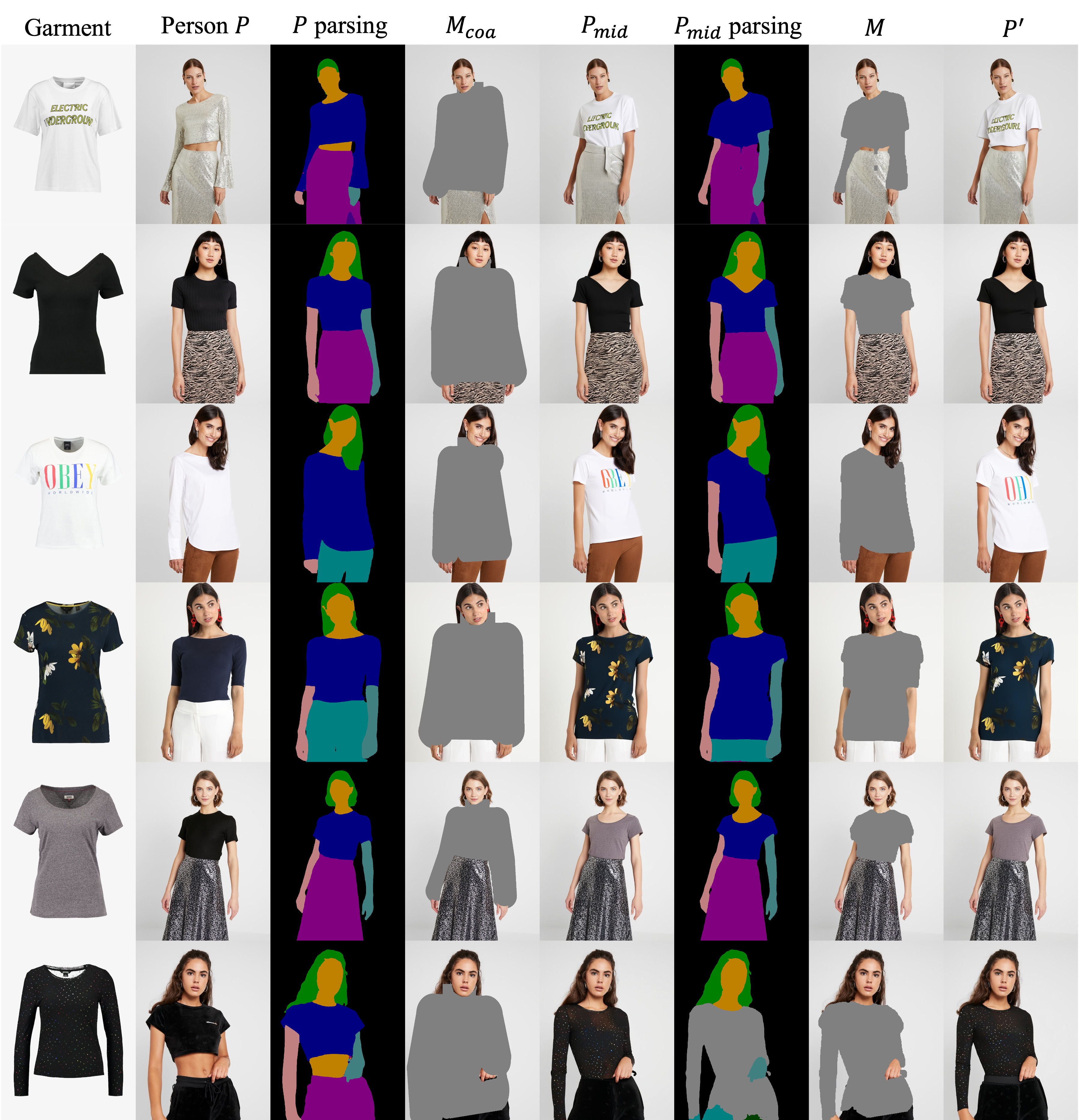}
    \caption{High-quality pseudo data generation in VITON-HD.
    \label{fig:HQ_hd}}
\end{figure*}

\begin{figure*}[h]
    \centering
    \includegraphics[width=0.8\textwidth]{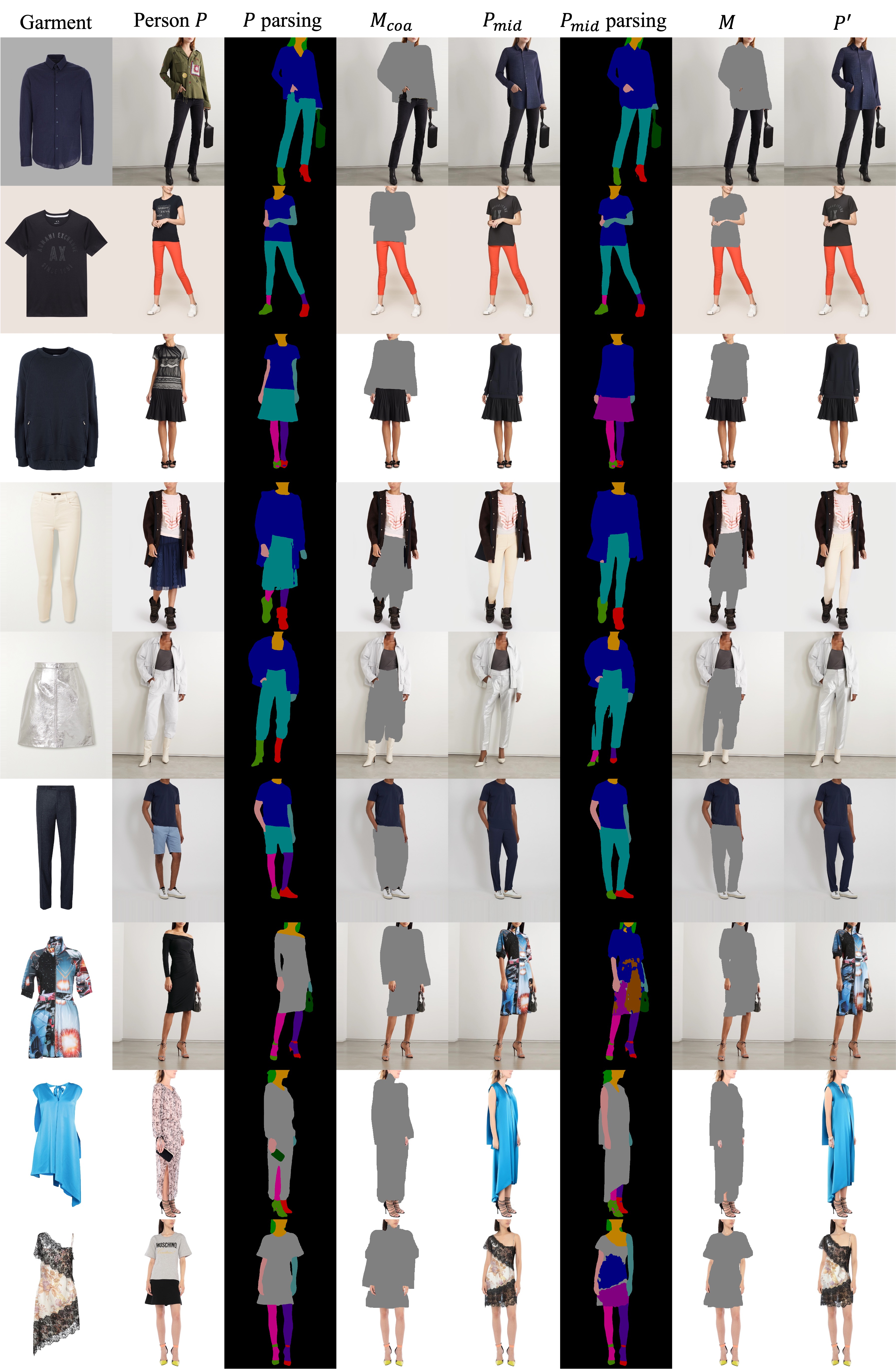}
    \caption{High-quality pseudo data generation in DressCode.
    \label{fig:HQ_dc}}
\end{figure*}

In this section, we provide additional details on creating High-Quality Pseudo Data. Figure \ref{fig:HQ_hd} and Figure \ref{fig:HQ_dc} illustrate the pseudo data used during training from the VITON-HD \cite{VITON-HD} and DressCode\cite{DressCode} datasets, respectively. After optimizing the masks, more information unrelated to the try-on process is preserved, reducing the negative impact of mask-based methods on the quality of pseudo data. The coarse mask $M_{coa}$ is generated using the same method as the masks in IDM-VTON \cite{IDM-VTON}.

\subsection{In-the-Wild Augmentation Details}
\begin{figure*}[h]
    \centering
    \includegraphics[width=0.5\textwidth]{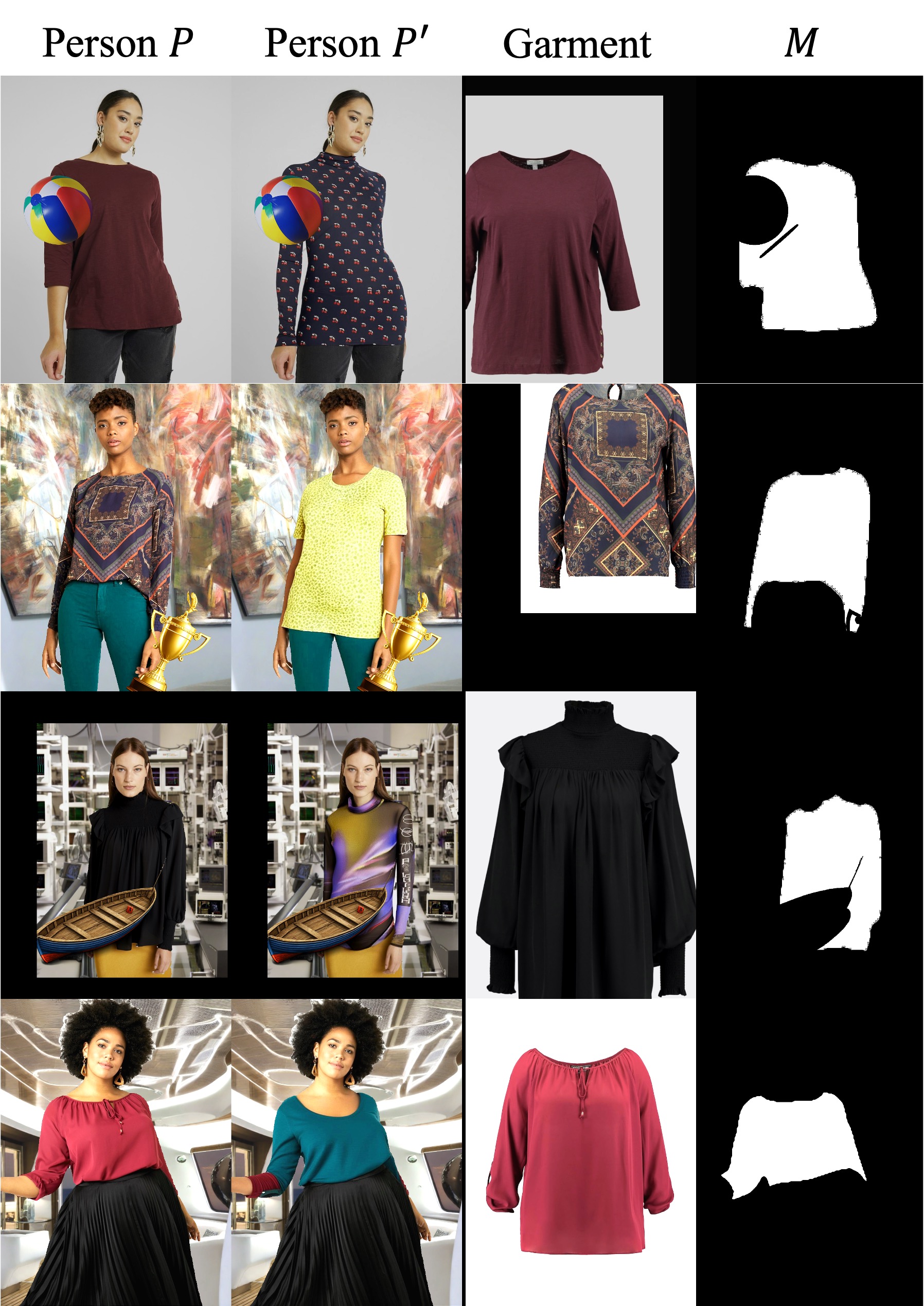}
    \caption{Data augmentation of training samples during the training process.
    \label{fig:wild_aug}}
\end{figure*}

\begin{figure*}[h]
    \centering
    \includegraphics[width=1\textwidth]{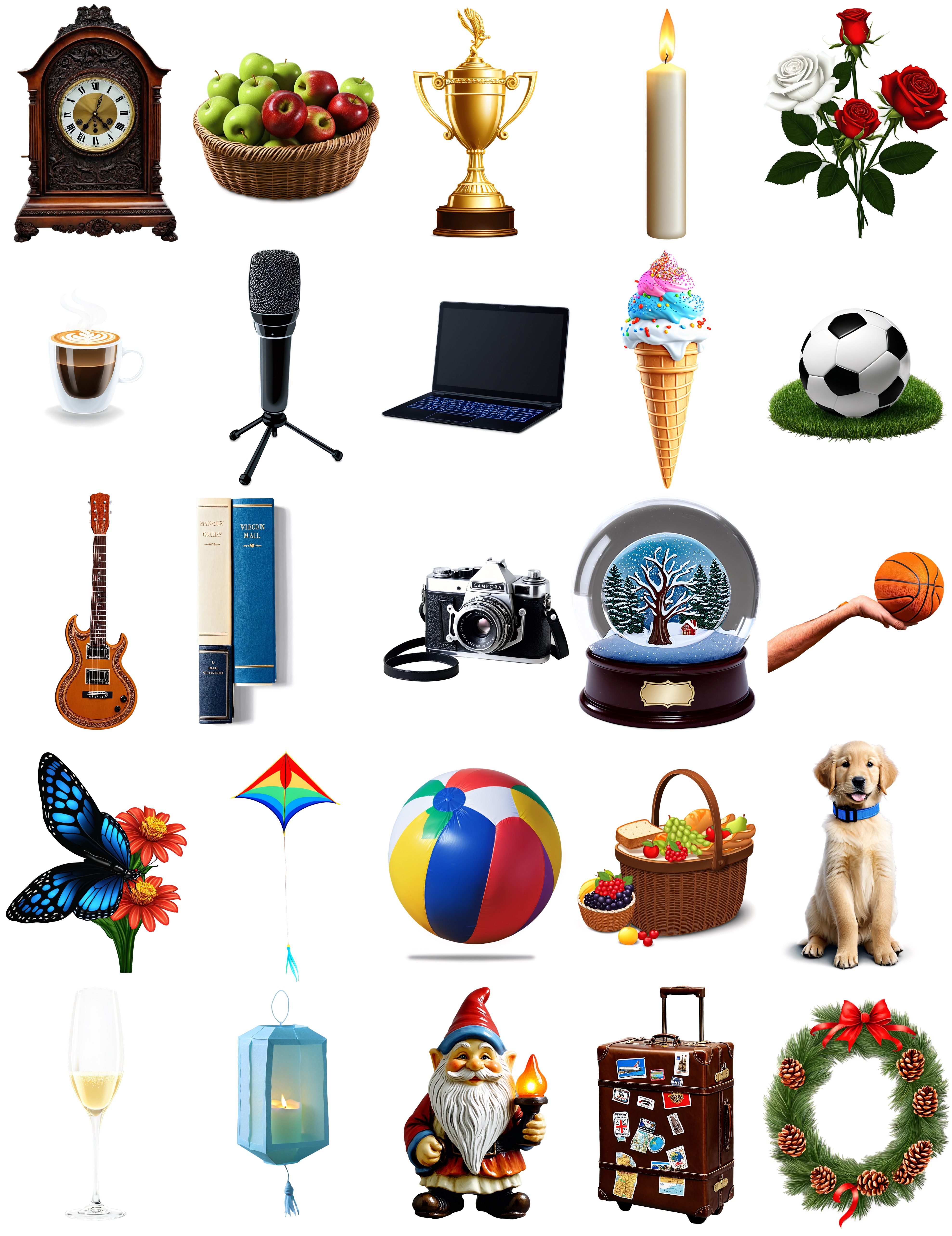}
    \caption{Foreground images used for in-the-wild data augmentation (Part 1).
    \label{fig:aug_fore_1}}
\end{figure*}

\begin{figure*}[h]
    \centering
    \includegraphics[width=1\textwidth]{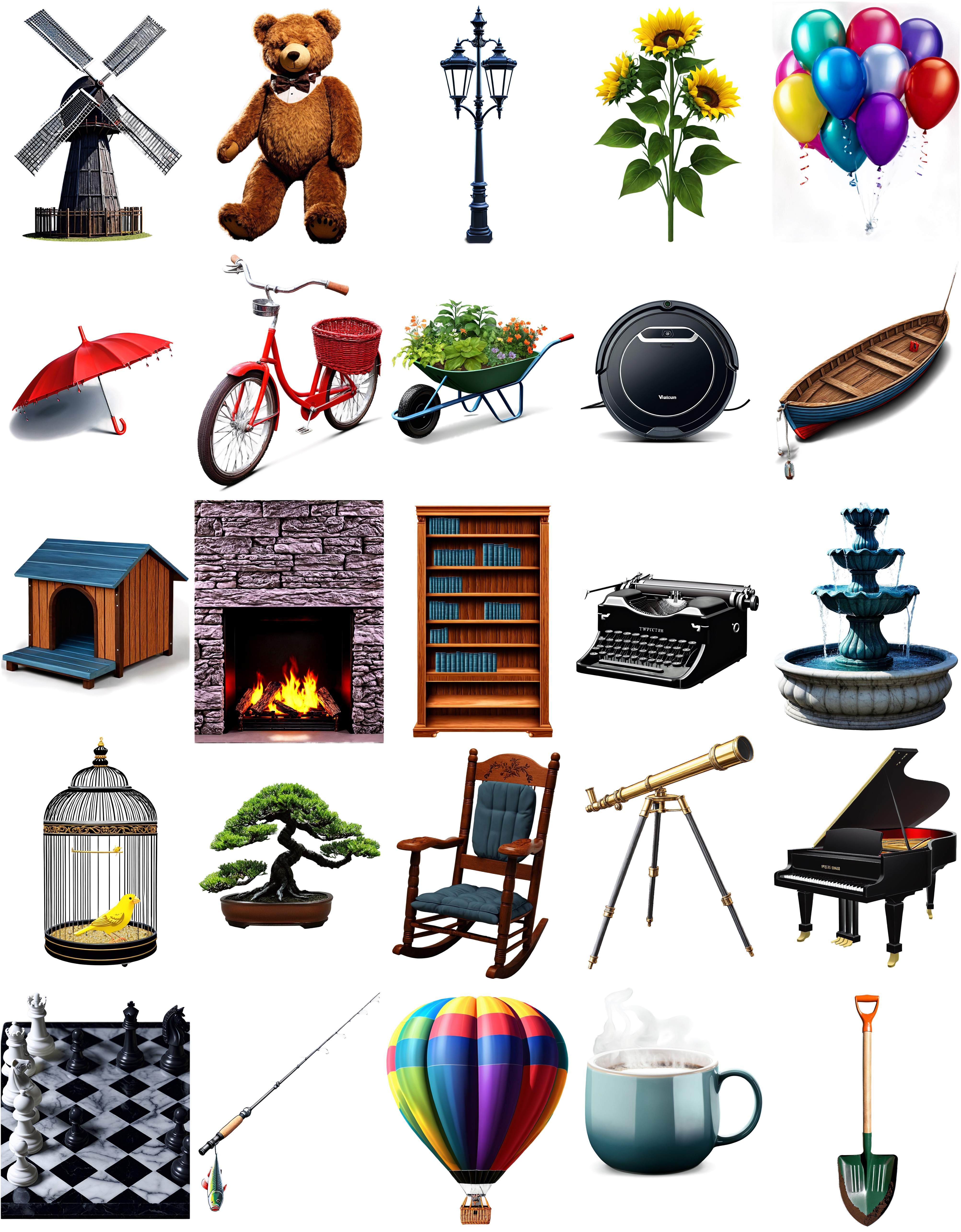}
    \caption{Foreground images used for in-the-wild data augmentation (Part 2).
    \label{fig:aug_fore_2}}
\end{figure*}

\begin{figure*}[h]
    \centering
    \includegraphics[width=1\textwidth]{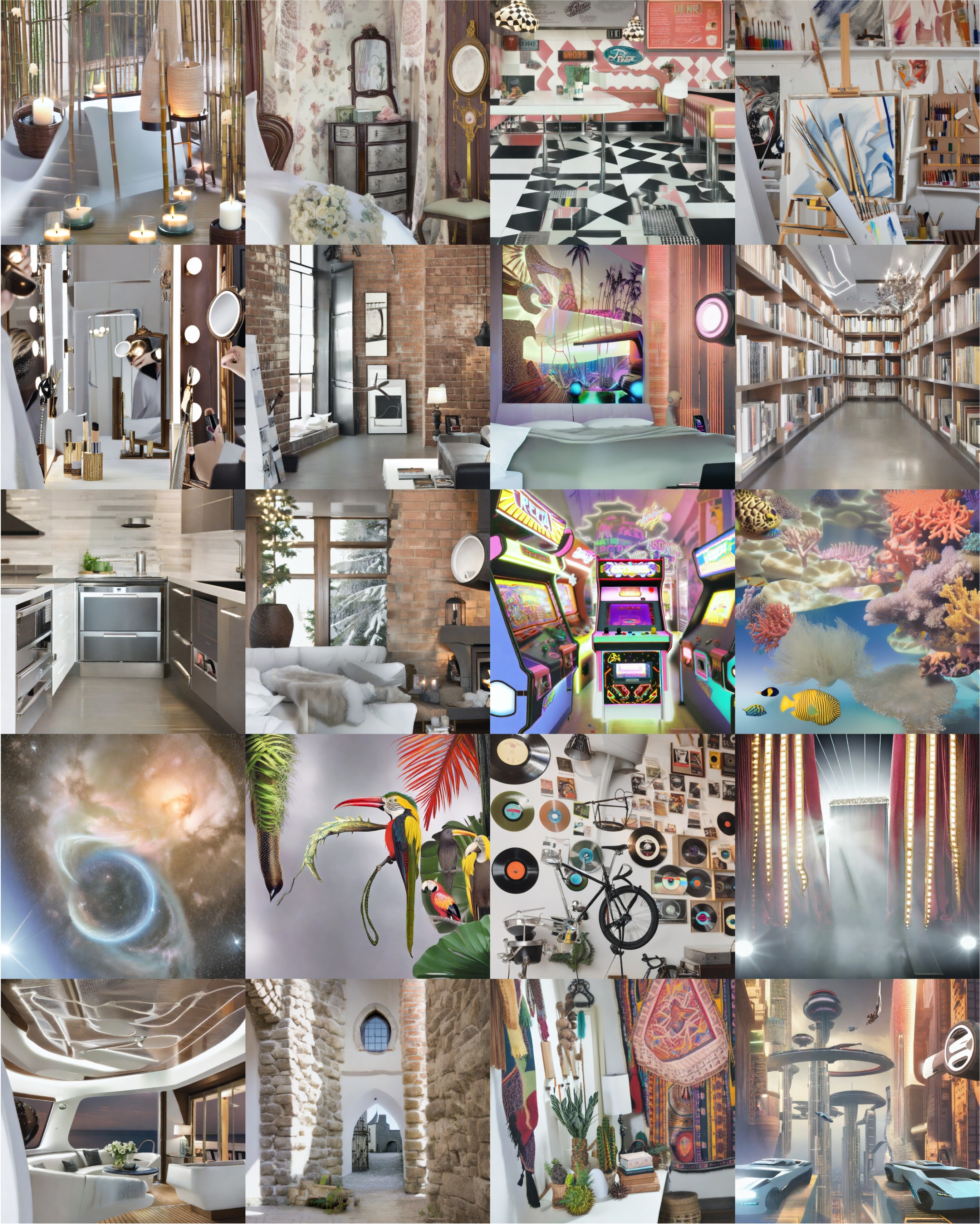}
    \caption{Background images used for in-the-wild data augmentation (Part 1).
    \label{fig:aug_back_1}}
\end{figure*}

\begin{figure*}[h]
    \centering
    \includegraphics[width=1\textwidth]{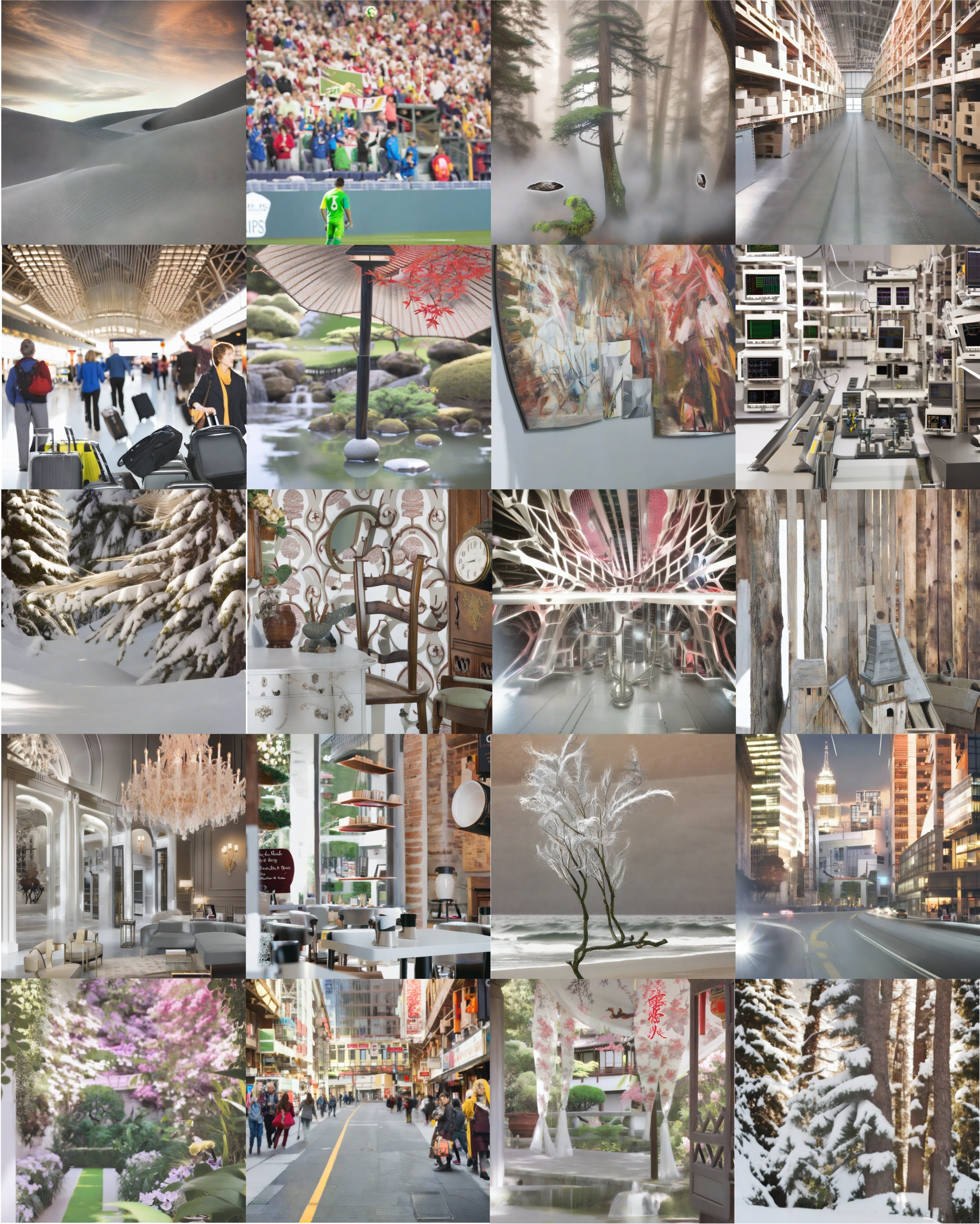}
    \caption{Background images used for in-the-wild data augmentation (Part 2).
    \label{fig:aug_back_2}}
\end{figure*}

In this section, we detail the implementation of wild data augmentation. In-the-Wild data augmentation utilizes 50 foreground images with a resolution of $1024\times768$, as shown in Figure \ref{fig:aug_fore_1}, Figure \ref{fig:aug_fore_2}, the corresponding prompts are shown in Table \ref{prompt_fore}. And 40 background images with a resolution of $1024\times1024$, as shown in Figure \ref{fig:aug_back_1}, Figure \ref{fig:aug_back_2}, the corresponding prompts are shown in Table \ref{prompt_back}. We independently add foregrounds and backgrounds to the person images. We add a background with a probability of p=0.5, where the placement of the person within the background image is chosen randomly. A foreground is applied with a probability of p=0.7, with the foreground image being randomly scaled (0.25 to 0.6 times) and randomly rotated (-45 to 45 degrees) before being overlaid on the person image. Based on in-the-wild data augmentation, we applied the same spatial data augmentation as StableVITON \cite{StableVITON}. As shown in Figure \ref{fig:wild_aug}, we provide visualization results of the training samples used during the training process.

\begin{table*}[h]
\centering
    \caption{Prompts for foreground images generation.}
    \begin{tabular}{c|p{0.7\linewidth}}
    \toprule
        Category  & Prompt \\
    \midrule
     \multirow{50}{*}{Foreground}& Colorful balloons, helium-filled, shiny and reflective\\
     & Sunflower bouquet, vibrant yellow, lush green stems\\
     & Vintage bicycle, red paint, woven wicker basket\\
     & Oversized teddy bear, brown fur, wearing a bow tie\\
     & Classic street lamp, wrought iron, glowing light\\
     & Red umbrella, open, raindrops dripping off edges\\
    & Christmas wreath, adorned with pinecones and red ribbon\\
    & Large suitcase, leather, covered in travel stickers\\
    & Garden gnome, ceramic, holding a small lantern\\
    & Floating lantern, paper, softly glowing candle inside\\
    & Sparkling wine glass, crystal clear, filled with champagne\\
    & Golden retriever puppy, playful, wearing a blue collar\\
    & Picnic basket, woven, filled with fruits and bread\\
    & Colorful kite, flying high, with long tail\\
    & Beach ball, multi-colored, bouncing in the air\\
    & Butterfly, vibrant wings, resting on a flower\\
    & Basketball, orange, spinning on a finger\\
    & Snow globe, intricate design, with a winter scene inside\\
    & Vintage camera, black and silver, with a leather strap\\
    & Book stack, old and new, with bookmarks sticking out\\
    & Guitar, wooden, with intricate inlays\\
    & Soccer ball, black and white, rolling on grass\\
    & Ice cream cone, double scoop, with sprinkles\\
    & Laptop, open, showing a bright screen\\
    & Microphone with stand, black, rich details\\
    & Cup of coffee, steaming, with latte art\\
    & Bouquet of roses, red and white, with green leaves\\
    & Candle, lit, with a gentle flame\\
    & Championship trophy, gold, rich details\\
    & Basket of apples, red and green, freshly picked\\
    & Antique clock, wooden frame, intricate carvings\\
    & Chessboard, marble, set up for a game\\
    & Fishing rod, graphite, with a lure attached\\
    & Hot air balloon, colorful, ascending into the sky\\
    & Coffee mug, ceramic, with steam rising\\
    & Garden shovel, metal, with soil on it\\
    & Piano, grand, with open lid\\
    & Telescope, brass, pointed towards the stars\\
    & Rocking chair, wooden, with cushions\\
    & Bonsai tree, miniature, with detailed branches\\
    & Birdcage, ornate, with a singing canary\\
    & Water fountain, marble, with flowing water\\
    & Typewriter, vintage, with paper inserted\\
    & Bookcase, wooden, filled with books\\
    & Fireplace, stone, with roaring fire\\
    & Dog house, wooden, with a nameplate\\
    & Fishing boat, wooden, anchored near the shore\\
    & Robot vacuum, sleek, cleaning the floor\\
    & Garden wheelbarrow, metal, filled with plants\\
    & Windmill, old-fashioned, turning in the wind \\
    \bottomrule
    \end{tabular}
\label{prompt_fore}
\end{table*}

\begin{table*}[h]
\centering
    \caption{Prompts for background images generation.}
    \begin{tabular}{c|p{0.7\linewidth}}
    \toprule
        Category  & Prompt \\
    \midrule
     \multirow{40}{*}{Background}& \{ \} in a china garden\\
    & \{ \} in a snowy winter landscape with pine trees\\
    & \{ \} in a bustling urban street scene\\
    & \{ \} in a garden, lush greenery\\
    & \{ \} in front of a modern city skyline\\
    & \{ \} on a sandy beach with ocean waves in the background\\
    & \{ \} in a cozy cafe setting with coffee cups and books\\
    & \{ \} in an elegant, upscale interior with chandeliers\\
    & \{ \} in a rustic countryside setting with fields and barns\\
    & \{ \} in a futuristic environment with metallic structures\\
    & \{ \} in a vintage-inspired room with old furniture and antiques\\
    & \{ \} in a snowy winter landscape with pine trees\\
    & \{ \} in a high-tech laboratory setting with monitors and equipment\\
    & \{ \} in a chic, modern art gallery\\
    & \{ \} in a traditional Japanese garden with pagodas and ponds\\
    & \{ \} in a busy airport terminal with travelers and luggage\\
    & \{ \} in an industrial warehouse with machinery and crates\\
    & \{ \} in a mystical forest setting with fog and ancient trees\\
    & \{ \} in a sports stadium during a match\\
    & \{ \} in a dramatic desert landscape with sand dunes and sunset\\
    & \{ \} in a futuristic cityscape with flying cars\\
    & \{ \} in a bohemian-style room with colorful tapestries\\
    & \{ \} in a medieval castle courtyard with stone walls\\
    & \{ \} in a luxury yacht setting with ocean views\\
    & \{ \} in a classical theater with velvet curtains and stage lights\\
    & \{ \} in a hipster cafe filled with vintage bicycles and records\\
    & \{ \} in a tropical rainforest with exotic plants and wildlife\\
    & \{ \} in a celestial-themed setting with stars and galaxies\\
    & \{ \} in an underwater scene with coral reefs and tropical fish\\
    & \{ \} in a 1980s retro arcade with neon lights and arcade games\\
    & \{ \} in a cozy living room with a fireplace and soft blankets\\
    & \{ \} in a sleek modern kitchen with stainless steel appliances\\
    & \{ \} in a sophisticated library surrounded by shelves of books\\
    & \{ \} in a futuristic bedroom with high-tech gadgets and mood lighting\\
    & \{ \} in a chic urban loft with exposed brick walls and industrial decor\\
    & \{ \} in a glamorous dressing room with mirrors and vanity lights\\
    & \{ \} in a minimalist art studio with canvases and paintbrushes\\
    & \{ \} in a retro diner with vinyl booths and checkerboard floors\\
    & \{ \} in a vintage-themed boudoir with lace curtains and antique furniture\\
    & \{ \} in a spa setting with candles, bamboo, and relaxation areas\\
    \bottomrule
    \end{tabular}
\label{prompt_back}
\end{table*}
\section{Ablation Study} 
\label{ablation}

\subsection{Additional Ablation Study Results}

\begin{figure*}[h]
    \centering
    \includegraphics[width=\textwidth]{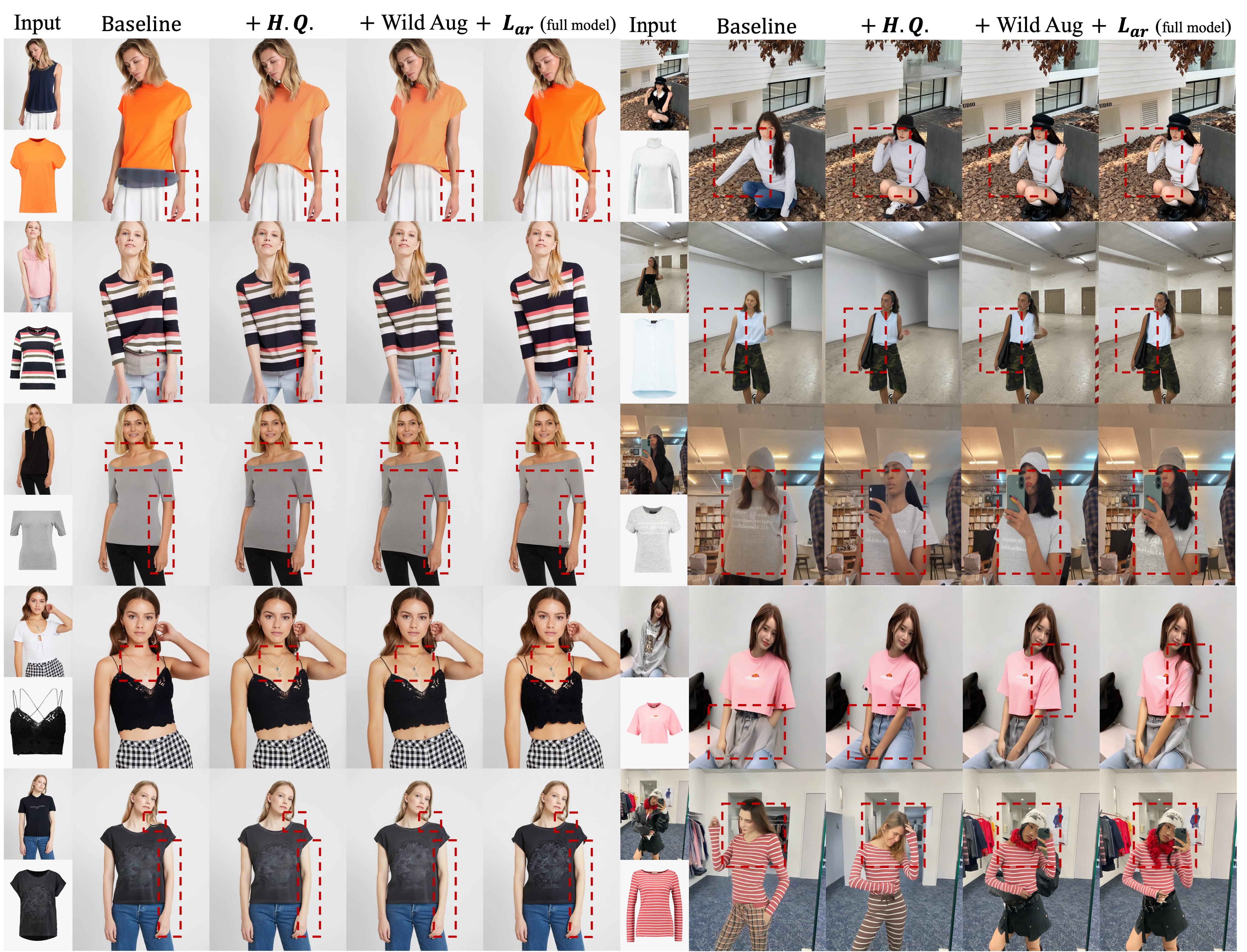}
    \caption{\textbf{Left}: Ablation study on the VITON-HD test set shows that H.Q. pseudo data enhances the retention of skin details and accessories.
\textbf{Right}: Ablation study on the WildVTON test set demonstrates that in-the-wild data augmentation significantly improves the retention of both foreground and background elements.
    \label{fig:ablation_st}}
\end{figure*}

\subsection{Study on the Effectiveness of Attention Layers}

\begin{figure*}[h]
    \centering
    \includegraphics[width=\textwidth]{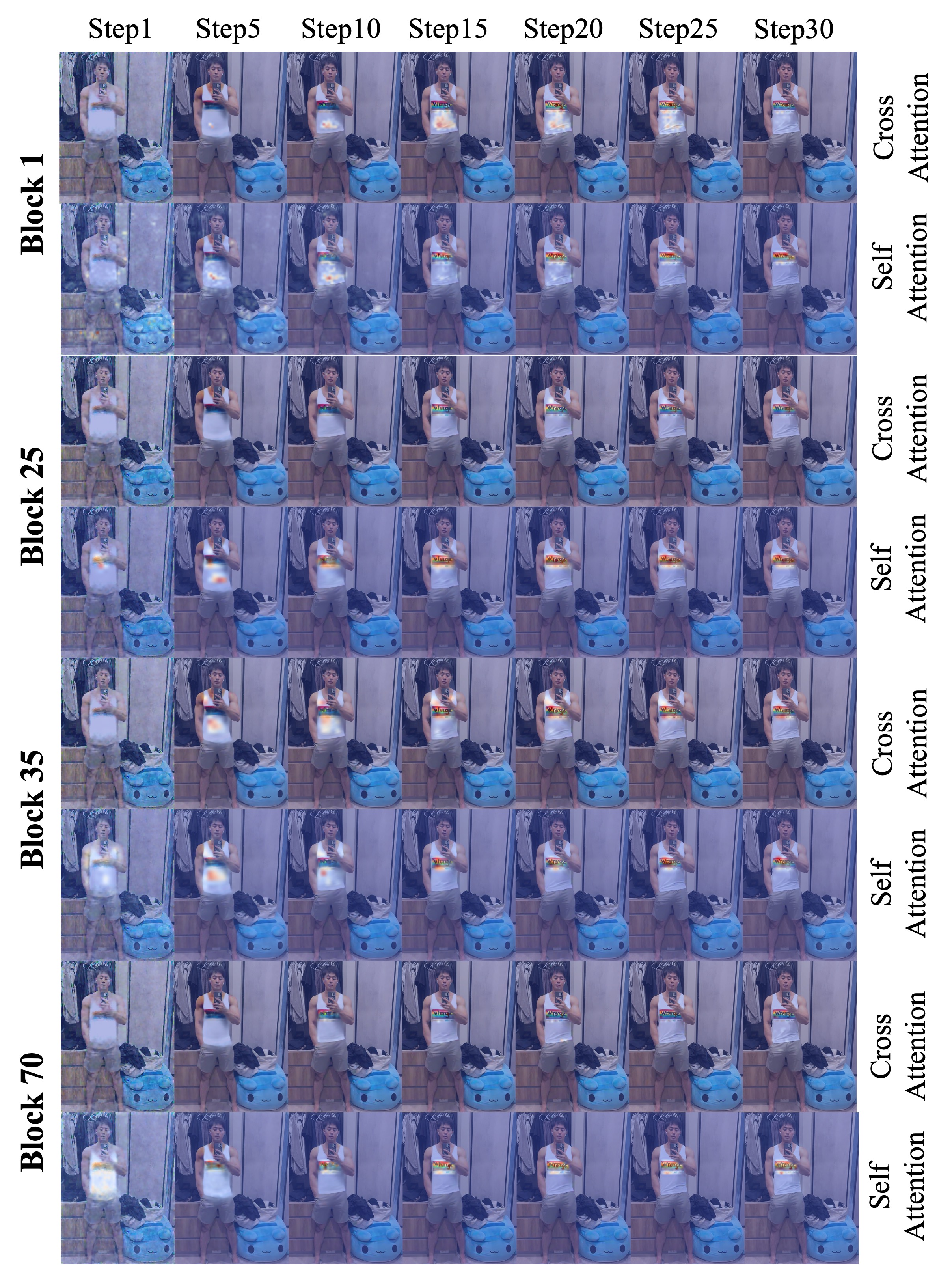}
    \caption{Visualization of attention heatmaps across different denoising stages and attention layers.
    \label{fig:ablation_layer_attn}}
\end{figure*}

Figure \ref{fig:ablation_layer_attn} presents an ablation study of attention layers in BooW-VTON. We selected steps 1, 5, 10, 15, 20, 25, and 30 from the DDIM sampler for analysis. We examined the cross-attention and self-attention layers from attention blocks 1, 25, 35, and 70 for each denoising step. As shown in Figure \ref{fig:ablation_layer_attn}, the model focuses on the distribution of clothing in the initial denoising stages, while focusing on the details of the garment in later stages. In the network, the attention layers near the middle stages exhibit greater activity.

\subsection{Attention Visualization in Ablation Study}
\begin{figure*}[h]
    \centering
    \includegraphics[width=\textwidth]{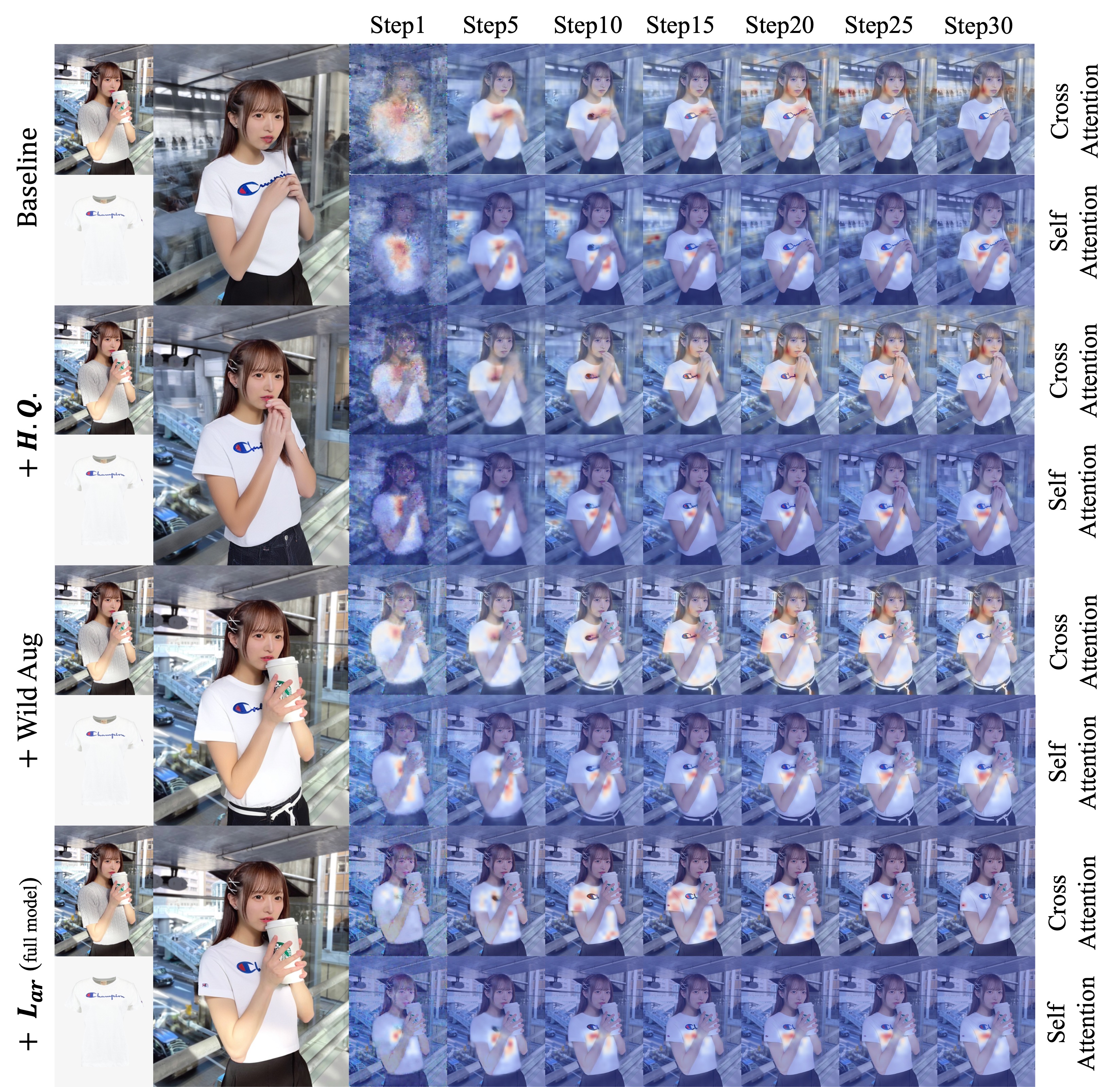}
    \caption{Cross-attention maps and self-attention maps in ablation experiments.
    \label{fig:ablation_attn_ab}}
\end{figure*}

We conducted ablation studies on the modules by visualizing the attention maps. As shown in Figure \ref{fig:ablation_attn_ab}, directly applying a mask-free approach fails to guide the model to focus on the correct try-on regions during the wild try-on, resulting in the scattering of attention across the entire image. After incorporating in-the-wild data augmentation, the model learns to retain content in non-try-on regions, shifting dispersed attention toward the try-on areas. With the application of the try-on localization loss, the attention maps are explicitly regularized, enabling the model to identify the try-on regions more accurately.

\section{Dataset Information}
\label{dataset}

\begin{figure*}[h]
    \centering
    \includegraphics[width=0.8\textwidth]{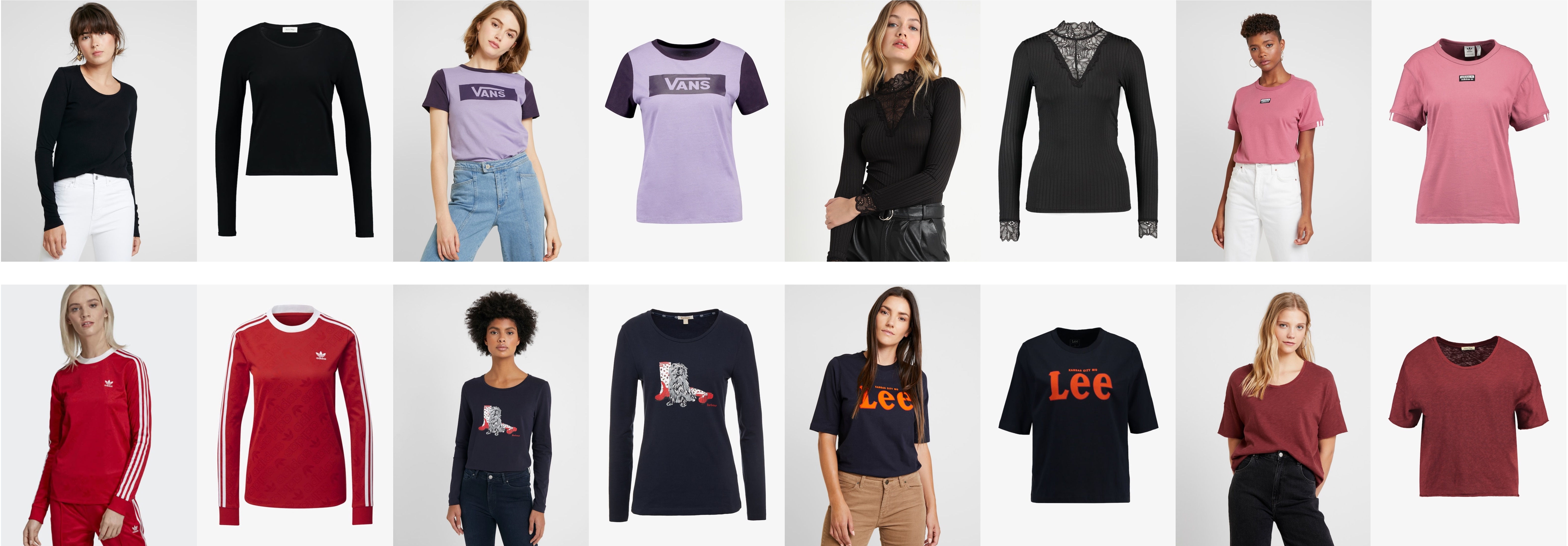}
    \caption{Samples of VITON-HD.
    \label{fig:dataset_hd}}
\end{figure*}

\begin{figure*}[h]
    \centering
    \includegraphics[width=0.8\textwidth]{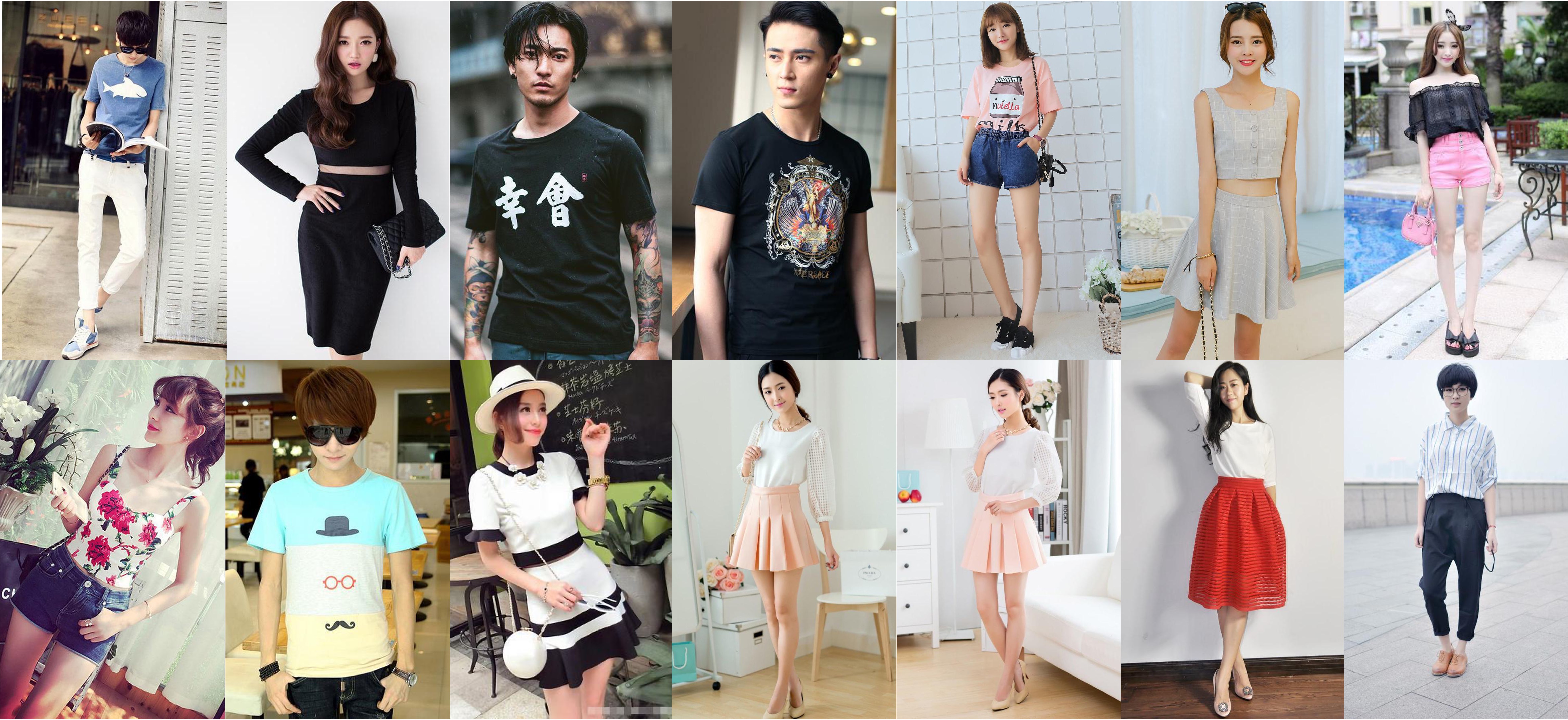}
    \caption{Samples of StreetVTON.
    \label{fig:dataset_st}}
\end{figure*}

\begin{figure*}[h]
    \centering
    \includegraphics[width=0.8\textwidth]{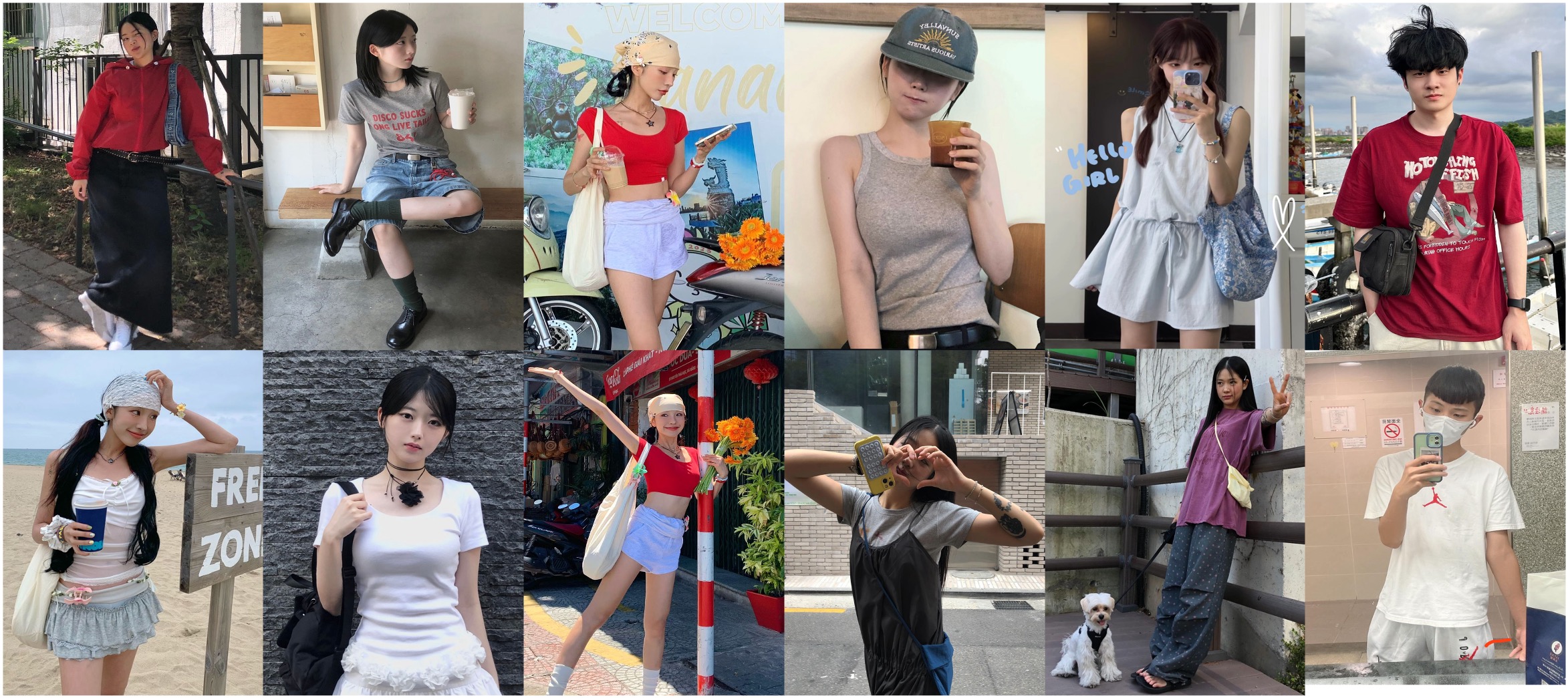}
    \caption{Samples of WildVTON.
    \label{fig:dataset_wd}}
\end{figure*}

\begin{figure*}[h]
    \centering
    \includegraphics[width=0.8\textwidth]{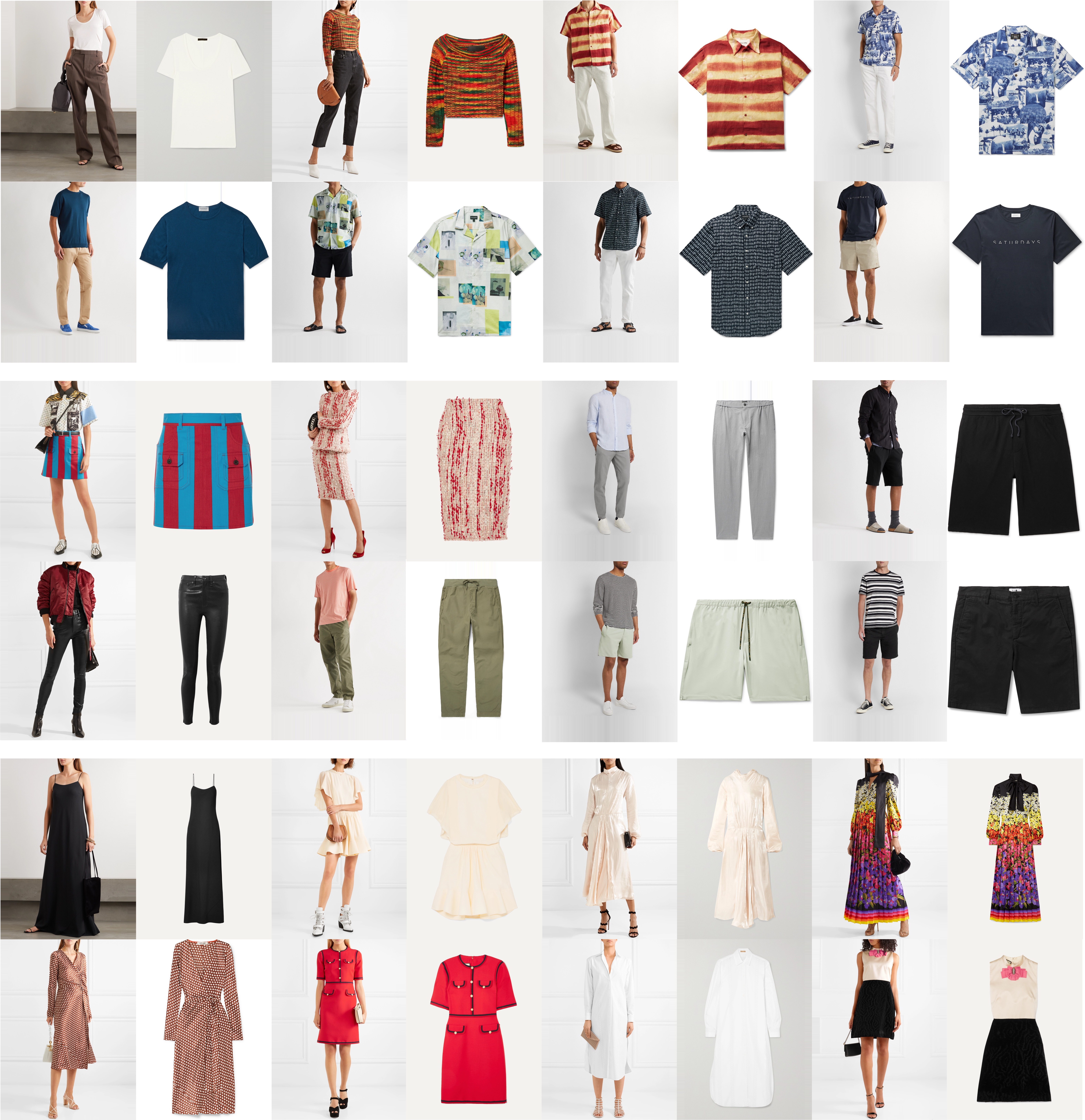}
    \caption{Samples of DressCode.
    \label{fig:dataset_dc}}
\end{figure*}

This section provides detailed information and image examples of the datasets used in the main text. 
As shown in \ref{fig:dataset_hd}, the VITON-HD \cite{VITON-HD} dataset consists of samples of women's spring and summer tops, with the training set containing 11,647 pairs and the test set containing 2,032 pairs. The DressCode \cite{DressCode} dataset is divided into three garment types: tops, bottoms, and dresses. It includes 29,478 dress samples, 15,363 top samples, and 8,951 bottom samples, with 1,800 pairs from each category extracted for the test set, as shown in \ref{fig:dataset_dc}. StreetVTON \cite{street_tryon} is a subset selected from DeepFashion2 \cite{deepfashion2} for in-the-wild try-on, as shown in \ref{fig:dataset_st}, containing 2,089 complex person images. WildVTON consists of 1224 lifestyle portrait images with complex foregrounds and backgrounds that we collected from the internet, as shown in \ref{fig:dataset_wd}. In addition, we validated the in-shop try-on performance on the VITON-HD and all three garment type DressCode test sets.


\end{document}